\documentclass[11pt,letterpaper]{iffyuan}

\usepackage[all]{hypcap}
\usepackage{algorithm}
\usepackage{algorithmicx}
\usepackage{algpseudocode}
\usepackage{float}
\usepackage{mathtools}
\usepackage{nicefrac}
\usepackage[caption=false,font=footnotesize]{subfig}
\usepackage{wrapfig}
\usepackage{xspace}
\usepackage{tikz}
\usepackage{array}
\usepackage{makecell}
\usepackage{stfloats}

\newcommand{\tele}{RoboPilot\xspace}
\newcommand{\rob}{AhaRobot\xspace}
\newcommand*\circled[1]{\tikz[baseline=(char.base)]{\node[shape=circle,fill=black,text=white,draw,inner sep=.1pt] (char) {#1};}}
\definecolor{lb}{rgb}{0.9, 0.95, 1.0}
\definecolor{purple}{rgb}{0.3, 0.1, 0.4}
\definecolor{softred}{rgb}{0.8, 0.2, 0.2}
\definecolor{grassgreen}{rgb}{0.2, 0.6, 0.2}
\newcommand{\down}{\color{softred}{$\downarrow$}\xspace}

\title{AhaRobot: A Low-Cost Open-Source Bimanual Mobile Manipulator for Embodied AI}

\author[1,*]{Haiqin Cui}
\author[1,*]{Yifu Yuan}
\author[1]{Yan Zheng}
\author[1,\textdagger]{Jianye Hao}

\affiliation[1]{Tianjin University}
\contribution[*]{Equal contribution}
\contribution[\textdagger]{Corresponding author}

\setteamlogo{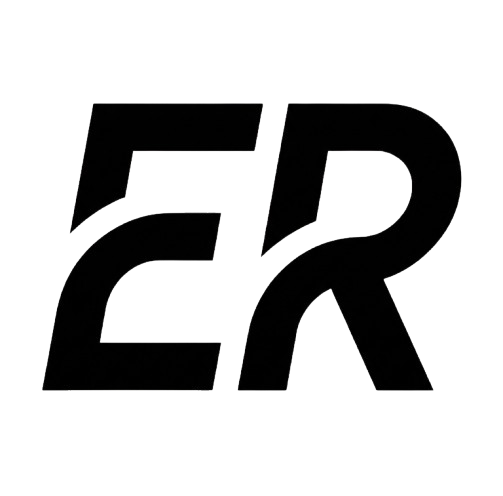}

\abstract{%
Scaling Vision-Language-Action models for embodied manipulation demands large volumes of diverse manipulation data, yet the high cost of commercial mobile manipulators and teleoperation interfaces that are difficult to deploy at scale remain key bottlenecks.
We present \textbf{\textit{\rob}}, a low-cost, fully open-source bimanual mobile manipulator tailored for Embodied-AI.
The system contributes: (1) a SCARA-like dual-arm hardware design that reduces motor torque demands while maintaining a large vertical reachable workspace, (2) an optimized control stack that improves precision via dual-motor backlash mitigation and static-friction compensation through dithering, and (3) \textbf{\textit{\tele}}, a teleoperation interface featuring a novel 26-faced marker handle for precise, long-horizon remote data collection.
Experimental results show that our hardware-control co-design achieves 0.7\,mm repeatability at a total hardware cost of only \$1,000.
The proposed 26-faced handle reduces tracking error by 80\% over a 6-faced baseline and improves data-collection efficiency by 30\%, while robustly handling singularities and supporting extremely long-horizon tasks in fully remote settings.
Despite its low cost, \rob enables imitation learning of complex household behaviors involving bimanual coordination, upper-body mobility, and contact-rich interaction, with data quality comparable to VR-based collection.
All software, CAD files, and documentation are available at \href{https://aha-robot.github.io}{https://aha-robot.github.io}.%
}

\metadata[\faGlobe\ Project]{\href{https://aha-robot.github.io}{https://aha-robot.github.io}}
\metadata[\faEnvelope\ Contact]{\href{mailto:cuihaiqin@tju.edu.cn}{cuihaiqin@tju.edu.cn}}

\begin{document}

\maketitle

\section{Introduction}

Recent advances in robotic manipulation~\cite{zhaoLearningFineGrainedBimanual2023, chiDiffusionPolicyVisuomotor2023, kimOpenVLAOpenSourceVisionLanguageAction2025, ghoshOctoOpenSourceGeneralist2024, liuRDT1BDIFFUSIONFOUNDATION2025} and navigation~\cite{liuDemonstratingOKRobotWhat2024, zhangNaVidVideobasedVLM2024, zhouNavGPT2UnleashingNavigational2025} have shown significant progress in embodied AI. Many tasks in everyday environments, such as cooking and house cleaning, require coordination of the entire body and dexterous use of dual arms. Therefore, bimanual mobile manipulators have been widely applied in embodied tasks~\cite{fuMobileALOHALearning2024, honerkampWholeBodyTeleoperationMobile2025, kempDesignStretchCompact2022, pollenReachy2023}. However, previous research hardware faced two issues: \textit{limited operational space} and \textit{affordability}.

Industrial-grade bimanual mobile manipulation platforms typically cost \textit{\$30,000} or more~\cite{palTiago2015, huangOpenPyRoA1Open2026, khazatskyDROIDLargeScaleInTheWild2024}, putting them out of reach for most laboratories. Lower-cost alternatives, in turn, sacrifice either precision or usable workspace~\cite{anjariaYORYourOwn2026, wang2025xlerobot, erciyesConeEOpenSource2025, pollenReachy2023}, making them inadequate for fine-grained household tasks. Existing systems thus present a clear trade-off: compact, affordable designs offer limited manipulation range, whereas fully capable platforms demand substantially higher cost and integration complexity~\cite{kempDesignStretchCompact2022, fuMobileALOHALearning2024, khazatskyDROIDLargeScaleInTheWild2024, pollenReachy2023}. This gap motivates the hardware design goal of this work: unifying dual-arm coordination, base mobility, and floor-level operation in a single platform that retains the precision and workspace required for fine household manipulation, while keeping cost low enough for large-scale deployment.

Using bimanual mobile robots for Imitation Learning~(IL) places demanding requirements on teleoperation data collection: operators must simultaneously control both arms and the mobile base for complex, long-horizon tasks across diverse scenarios, far exceeding the complexity of desktop manipulation. VR-based approaches~\cite{chengOpenTeleVisionTeleoperationImmersive2024, dingBunnyVisionProRealTimeBimanual2024, heOmniH2OUniversalDexterous2024} can capture end-effector poses, but the bulky headsets are unsuitable for prolonged remote sessions, and additional base-controller adaptation is needed for full mobile-robot control. 3D mice~\cite{luoPreciseDexterousRobotic2024} or joysticks offer another alternative, yet they struggle with whole-body remote operation. Exoskeleton-like leader-follower schemes~\cite{fuMobileALOHALearning2024, zhaoLearningFineGrainedBimanual2023, wuGELLOGeneralLowCost2023, fangAirExoLowCostExoskeletons2024, yangACECrossplatformVisualExoskeletons2024, shawBimanualDexterityComplex2024} construct a kinematically matched leader arm to capture joint-angle data, but require extra motors or encoders; moreover, representative systems such as Mobile Aloha demand on-site operation, lack remote-control capability, and scale poorly to long-distance tasks.

\begin{figure}[!htbp]
    \centering
    \includegraphics[width=1\linewidth]{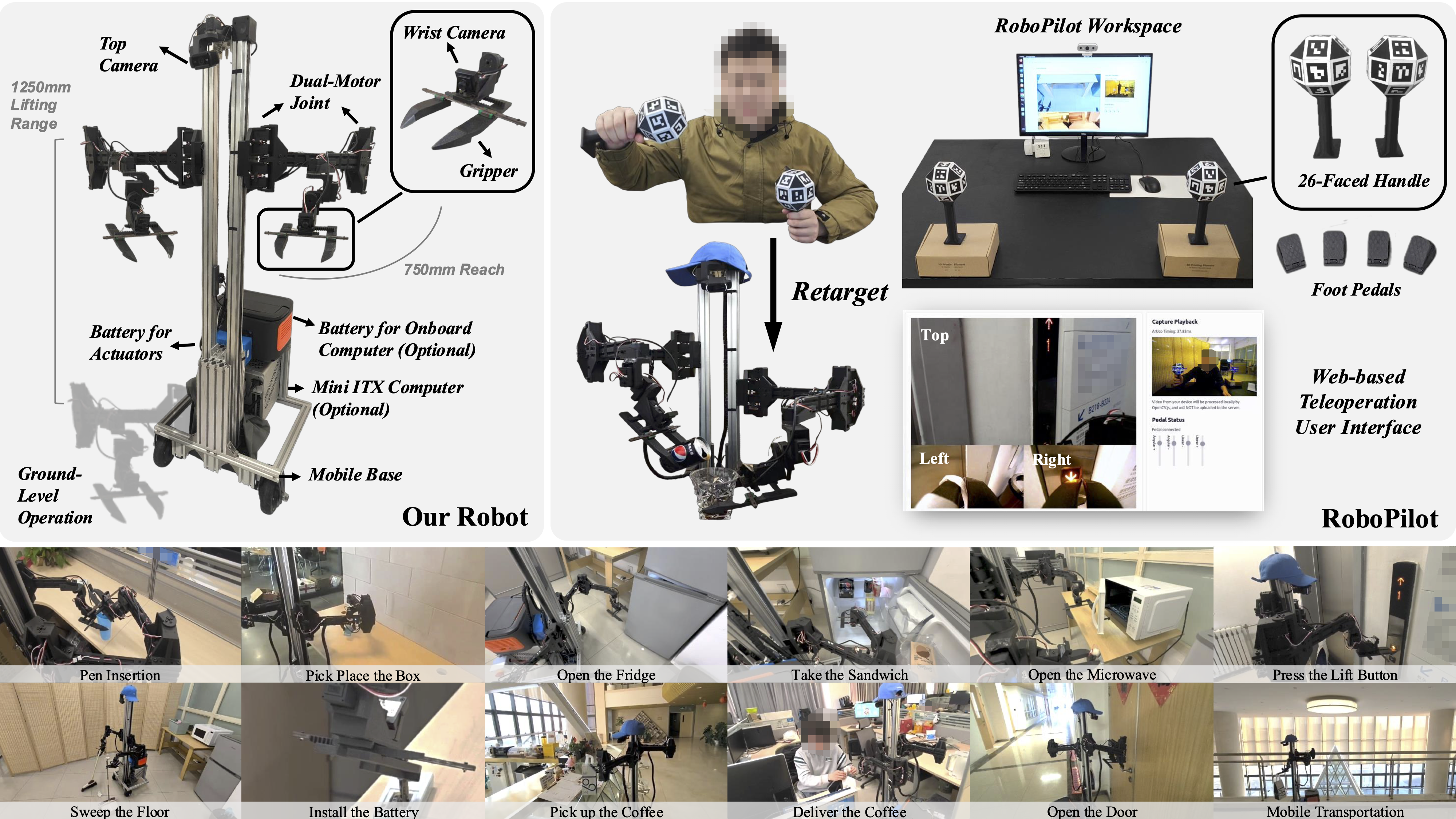}
    \caption{\textbf{Overview of \rob.} \textit{Top left:} Hardware design: lifting rails provide a 1250\,mm vertical stroke with floor-reaching capability, and SCARA-like horizontal dual arms with dual-motor joints improve end-effector precision. \textit{Top right:} The proposed teleoperation method \tele, comprising novel 26-faced marker handles, foot pedals, and a web-based full-body teleoperation interface. Hand pose is captured by a webcam on the operator's computer and retargeted to the robot in real time. \textit{Bottom:} Examples of diverse tasks performed by \rob in real home environments, spanning fine-grained manipulation, bimanual coordination, whole-body mobility, and floor-level interaction.}
    \label{fig:main}
\end{figure}

To address these challenges, we introduce \textbf{\textit{\rob}}, a low-cost open-source bimanual mobile manipulator with a base hardware budget of \textit{\$1,000}. As shown in~\cref{tab:robot_comparison}, \rob offers comprehensive functionality and workspace at roughly 1/15 the cost of popular platforms (system overview in~\cref{fig:main}). The system comprises three components: \circled{1} \textit{Hardware Configuration} built entirely from off-the-shelf parts, providing dual-arm manipulation, whole-body mobility, and floor-reaching capability for remote failure recovery; \circled{2} \textit{Control System} with dual-motor anti-backlash and dithering-based friction compensation for improved tracking precision; \circled{3} \textit{Teleoperation} via \textbf{\textit{\tele}}, which combines a 26-faced marker handle and foot pedals for intuitive full-body control, reducing marker-based tracking error by 80\% and achieving data quality competitive with VR-based collection. \tele requires only \textit{\$50} (a webcam and 3D-printed parts), supports fully remote operation, and enables scalable crowdsourced data collection. Experiments spanning control precision, marker-handle tracking stability, long-horizon remote task execution, teleoperation efficiency, and imitation learning validate the practicality of \rob for real-world embodied-AI applications.

Our contributions are summarized as follows.
\begin{itemize}
\item \textbf{Hardware Design:} We present a human-scale bimanual mobile manipulator built entirely from off-the-shelf parts, featuring a lifting mechanism and counter-gravity joint arrangement that expands the vertical workspace while reducing required joint torque.
\item \textbf{Control Optimization:} With dual-motor backlash mitigation and dithering-based friction compensation, low-cost components achieve 0.7\,mm repeatability and reliably track incremental targets down to two encoder-resolution steps.
\item \textbf{\tele Teleoperation:} We introduce a 26-faced marker-handle teleoperation framework that reduces tracking error by 80\% with up to 1\,mm spatial resolution, costs only \textit{\$50}, supports fully remote operation, and delivers data quality competitive with VR-based collection.
\item \textbf{Open-Source Ecosystem:} We release complete CAD files, control software, and deployment documentation; following the provided guide, the entire system can be assembled from scratch in approximately three days.
\end{itemize}

\begin{table}[!htbp]
    \centering
    \caption{Comparison of Different Robotic Platform. $\dagger$: \$1,000 version does not include computing resources, while \$1,800 version is for the mini computer with RTX4060 GPU.}
    \setlength{\tabcolsep}{2pt}
    \renewcommand{\arraystretch}{1.1}
    \begin{tabular}{l|lccccc}
        \toprule
        \textbf{Robot Platform} & \textbf{Price} & \makecell{\textbf{Dual} \\ \textbf{Arm}} & \makecell{\textbf{Mobile} \\ \textbf{Base}} & \makecell{\textbf{Reach} \\ \textbf{Floor}} & \makecell{\textbf{Hardware} \\ \textbf{OpenSource}} & \textbf{DoF} \\
        \midrule
        Mobile Aloha~\cite{fuMobileALOHALearning2024} & \$32,000 & $\checkmark$ & $\checkmark$ & $\times$ & $\checkmark$ & 16 \\
        Hello Robot~\cite{kempDesignStretchCompact2022} & \$24,950 & $\times$ & $\checkmark$ & $\checkmark$ & $\times$ & 7 \\
        DROID~\cite{khazatskyDROIDLargeScaleInTheWild2024} & $\approx \$27,000$ & $\times$ & $\times$ & $\times$ & $\times$ & 8 \\
        AgileX COBOT & $\approx \$30,000$ & $\checkmark$ & $\checkmark$ & $\times$ & $\times$ & 16 \\
        Reachy~\cite{pollenReachy2023} & $\approx \$75,000$ & $\checkmark$ & $\checkmark$ & $\times$ & $\checkmark$ & 19 \\
        TIAGo~\cite{palTiago2015} & $>\$200,000$ & $\times$ & $\checkmark$ & $\checkmark$ & $\times$ & 12 \\
        OpenPyRo-A1~\cite{huangOpenPyRoA1Open2026} & \$14,000 & $\checkmark$ & $\times$ & $\times$ & $\checkmark$ & 16 \\
        XLeRobot~\cite{wang2025xlerobot} & \$660 & $\checkmark$ & $\checkmark$ & $\times$ & $\checkmark$ & 15 \\
        Cone-E~\cite{erciyesConeEOpenSource2025} & \$12,000 & $\checkmark$ & $\checkmark$ & $\checkmark$ & $\times$ & 19 \\
        \cellcolor{lb}$\textbf{Ours}^\dagger$ & \cellcolor{lb}\textbf{\$1,000-1,800} & \cellcolor{lb}$\checkmark$ & \cellcolor{lb}$\checkmark$ & \cellcolor{lb}$\checkmark$ & \cellcolor{lb}$\checkmark$ & \cellcolor{lb}\textbf{16} \\
        \bottomrule
    \end{tabular}
    \label{tab:robot_comparison}
\end{table}

\FloatBarrier
\section{Related Works}

\textbf{Mobile Manipulator:} A common approach to building a mobile manipulator is to mount an industrial robotic arm on a mobile base~\cite{xiongAdaptiveMobileManipulation2024, shawBimanualDexterityComplex2024, spahnDemonstratingAdaptiveMobile2024, fuMobileALOHALearning2024, pengRevolutionizingBatteryDisassembly2024, wuTidyBotOpenSourceHolonomic2024}. While effective for basic tasks, these platforms typically lack dual-arm coordination and vertical lifting capability, which are essential for whole-body household manipulation~\cite{fuMobileALOHALearning2024, anjariaYORYourOwn2026, wang2025xlerobot, erciyesConeEOpenSource2025}.
Some research~\cite{xieCoupledActivePerception2024, liDynamicInteractionControl2024, zhangLearningOpenTraverse2024} replaces the wheeled base with a quadrupedal robot to enhance terrain adaptability, yet these designs still struggle with limited lifting ability and dual-arm coordination, resulting in a constrained working area.
Another line of work~\cite{bajracharyaMobileManipulationSystem2020, bajracharyaDemonstratingMobileManipulation2023, smithDesignStickbugSixArmed2024, lenzNimbRoWinsANA2023, huangOpenPyRoA1Open2026} develops entirely new configurations from scratch. These typically involve multiple custom CNC parts that are difficult to fabricate and hinder reproducibility. Hello Robot Stretch~\cite{kempDesignStretchCompact2022} similarly relies on numerous custom components for its telescoping arm, leading to a complex structure and high cost.
In contrast, \rob is built entirely from off-the-shelf components with a lifting-rail mechanism that eliminates counter-gravity torque requirements, achieving high degrees of freedom and a large workspace at low cost. A complete open-source guide enables exact replication.

\textbf{Teleoperation:} Collecting scalable, high-quality demonstrations for mobile manipulation requires a teleoperation system that simultaneously addresses whole-body control (dual arms, mobile base, and vertical lifting), fully remote operation, and low-cost easy deployment.
VR-based approaches~\cite{chengOpenTeleVisionTeleoperationImmersive2024, dingBunnyVisionProRealTimeBimanual2024, heOmniH2OUniversalDexterous2024, iyerOPENTEACHVersatile2024} capture end-effector poses but require bulky headsets unsuitable for prolonged remote sessions, and need additional base-controller adaptation for full mobile-robot control.
Motion-capture-suit methods~\cite{setapenMARIOnETMotionAcquisition2010, stantonTeleoperationHumanoidRobot2012} demand significant hardware investment and are difficult to scale.
UMI family works~\cite{chiUniversalManipulationInterface2024, wuFastUMIScalableHardwareIndependent} combine a low-cost gripper with SLAM for handheld data collection; however, SLAM pipelines require substantial engineering effort for tuning and deployment, and may suffer from localization failures and high latency, making them less suitable for real-time remote operation.
Exoskeleton-like leader-follower systems~\cite{fangAirExoLowCostExoskeletons2024, yangACECrossplatformVisualExoskeletons2024, zhaoLearningFineGrainedBimanual2023, fuMobileALOHALearning2024, wuGELLOGeneralLowCost2023, shawBimanualDexterityComplex2024} construct kinematically matched leader arms to capture joint-angle data, but require additional electronic components and extensive manual assembly.
Moreover, most methods focus on upper-body teleoperation, leaving base control underexplored: some solutions~\cite{fuMobileALOHALearning2024} require the operator to physically push the robot, while others rely on a second person~\cite{wuGELLOGeneralLowCost2023} to drive the base.
\tele addresses these gaps with a 26-faced marker handle, foot pedals, and a web-based interface, achieving whole-body remote teleoperation at only \textit{\$50}. The lightweight setup is easy to deploy and imposes low operator burden, facilitating large-scale crowdsourced data collection.

\FloatBarrier
\section{Low-Cost Hardware System}

The hardware design of \rob follows a core principle: minimizing joint torque requirements through mechanical structure so that the entire system can be built from low-cost off-the-shelf components without sacrificing precision or workspace. Specifically, the design satisfies four requirements: 1) \textbf{Affordability:} joint configuration and component selection are optimized for cost efficiency; 2) \textbf{Whole-Body Mobility:} navigate different locations and execute tasks at various heights; 3) \textbf{Minimal Footprint:} compact design to pass through confined spaces; 4) \textbf{Without On-site Assistance:} recover from failures fully remotely. Core parameters are summarized in~\cref{tab:robot_parameters}.

\begin{table}[!htbp]
    \caption{\small Core Parameters of Proposed Robot.}
    \centering
    \begin{tabular}{ll}
        \toprule
        \textbf{Parameter}      & \textbf{Value}  \\
        \midrule
        Payload (Single Arm)    & 1.5 kg          \\
        Repeatability           & 0.7 mm          \\
        Weight                  & 51 kg           \\
        Size                    & $550\times500\times1550$ mm \\
        Max Gripper Width       & 120 mm          \\
        Max Reach (X-Y Plane)   & 750 mm          \\
        Z-Axis Reach            & 1250 mm         \\
        Battery Life            & 4-5 hr          \\
        Min Turning Radius      & 0               \\
        Turning Sweeping Radius & 500 mm          \\
        \bottomrule
    \end{tabular}
    \label{tab:robot_parameters}
\end{table}

\begin{figure}[!htbp]
    \centering
    \includegraphics[width=1\linewidth]{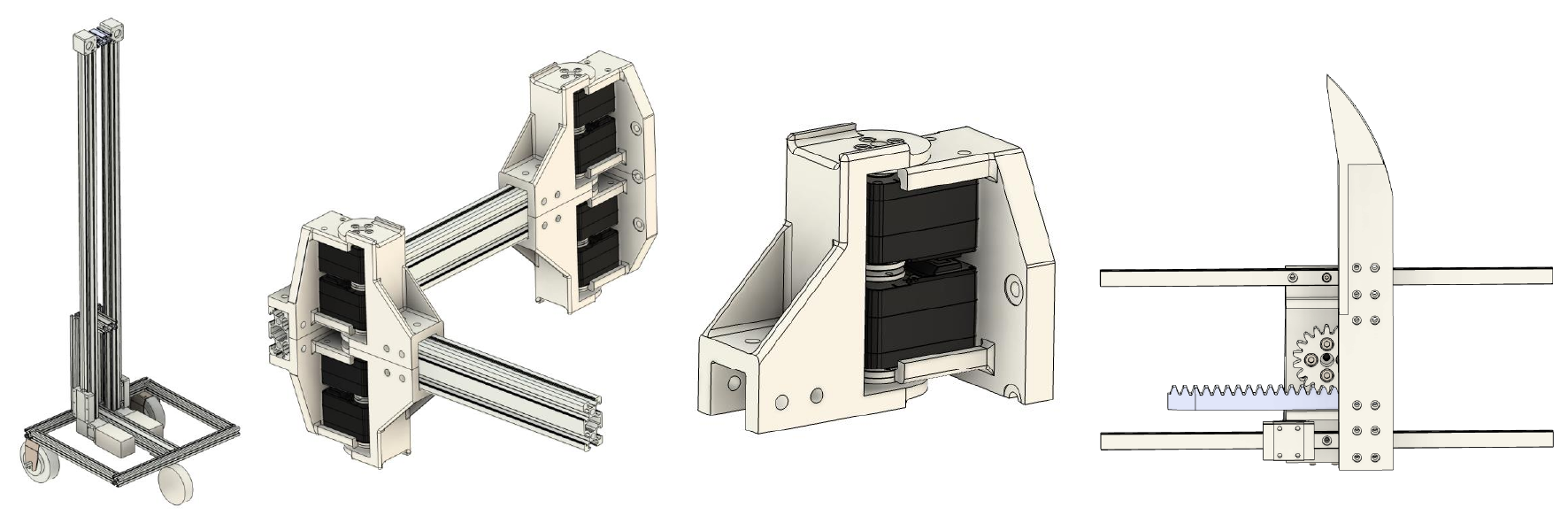}
    \caption{\textbf{Modular Mechanical Design of \rob.} From left to right: (a) two lifting rails that realize vertical arm motion while also serving as the upper-body structure; (b) a horizontally arranged arm architecture composed of four modules; (c) dual-motor joints; and (d) the gripper module. All components are designed in a modular manner. Screw holes are placed on the outer side to simplify maintenance, enabling rapid joint replacement in case of failure.}
    \label{fig:robot_modular_design}
    \vspace{-15pt}
\end{figure}

\subsection{Morphology}

As illustrated in~\cref{fig:main,fig:robot_modular_design}, \rob consists of three main modules: lifting rails that double as the upper-body structure, horizontally arranged SCARA-like dual arms, and a differential-drive mobile base. All structural parts are built from aluminum profiles and 3D-printed components in a modular fashion, enabling full assembly in approximately three days.

\textbf{Lifting Capability:} The absence of lifting freedom is a common pain point for mobile manipulators: in our early prototyping, failures such as objects slipping from the gripper occurred frequently and could not be recovered remotely, requiring on-site human intervention. To satisfy \textit{requirements~2} (multi-height operation) and \textit{4} (no on-site assistance), \rob integrates vertical lifting degrees of freedom. Lead screws offer high positioning accuracy and load capacity but are too slow for dynamic tasks; we therefore chose belt-driven slides that balance speed and cost, while letting the two rails double as the upper-body frame to eliminate extra structural parts (\textit{requirement~1}).

\textbf{SCARA-like Arms:} Conventional articulated arms must continuously counteract gravity at the shoulder and elbow, demanding expensive high-torque motors. Inspired by the SCARA configuration, we arrange the arms horizontally and delegate vertical motion to the lifting rails, thereby eliminating the counter-gravity load on all rotational joints---motors need only overcome arm inertia and payload. This is the key design choice that enables the use of low-cost Feetech STS3215 servos (\textit{requirement~1}) while still providing a 750\,mm horizontal reach. The horizontal layout also allows the arms to fold against the body during standby or transit, minimizing space occupation (\textit{requirement~3}).

\textbf{Dual-Motor Joints:} Low-cost servos (Feetech STS3215, brushed DC with 1:345 gearbox, 35\,kg$\cdot$cm max torque) are affordable (\textit{requirement~1}) but suffer from significant gear backlash. Drawing an analogy from antagonistic muscle pairs~\cite{pfeifer2007self}, we couple two motors per joint and apply opposing bias torques so that the gear teeth remain in constant contact, eliminating backlash. Each joint is designed as a self-contained module with externally accessible fasteners, enabling rapid replacement in case of failure. The control details of this dual-motor scheme are presented in~\cref{sec:dual-joint-control}.

\textbf{Differential Drive Mobile Base:} Two front-mounted BLDC motors and a rear castor wheel provide holonomic-like mobility with zero-radius turning (\textit{requirements~2 and 3}). The chassis is assembled from aluminum profiles---avoiding CNC machining or casting (\textit{requirement~1})---and its front-to-back length is minimized to limit the sweeping radius to 50\,cm, reducing collision risk in confined household spaces.

\subsection{Sensing, Computing, and Power}

\rob carries three webcams: one on a 2-DoF pan-tilt head gimbal for an adjustable panoramic view across different working heights, and two on the left and right wrists for close-range grasp observation. A resolution of $640\times360$ at 30\,Hz is chosen to balance image quality with WebRTC streaming bandwidth during remote teleoperation. A photoelectric switch at the base of the lifting slider provides homing after power-on or stepper step-loss events.

For on-board inference of end-to-end policies, \rob is equipped with a Mini-ITX computer (Intel i5-12700KF CPU, NVIDIA RTX4060 GPU). Low-level control is distributed across five ESP32 microcontrollers that handle motion profiling and PID for the arms, head, and lift, plus an ODrive 3.6 for the base BLDC motors; all modules communicate via ROS\,2 Humble. Actuators are powered by a 24\,V / 20\,Ah (294\,Wh) lithium-polymer battery, while a 1\,kWh portable AC supply powers the computer, yielding 4--5 hours of untethered operation. Both computing and power modules are rear-mounted to balance the center of gravity. An emergency-stop switch provides immediate system shutdown.

\FloatBarrier
\section{Dual-Joint Control}
\label{sec:dual-joint-control}

\begin{figure}[!htbp]
    \centering
    \includegraphics[width=0.8\linewidth]{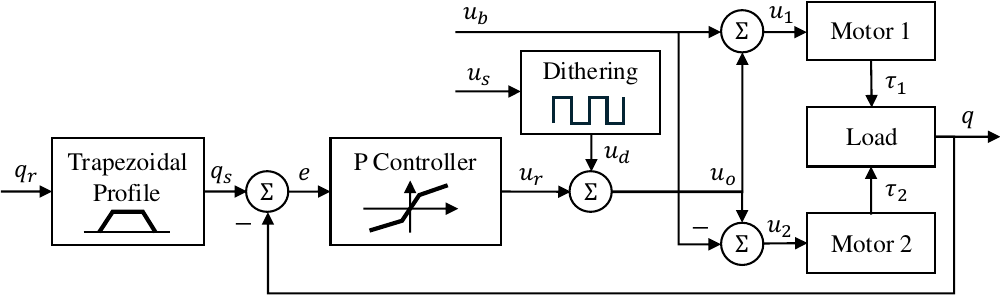}
    \caption{\textbf{Block Diagram of Dual-Joint Control System.} In addition to trapezoidal motion profiles that smooth acceleration/deceleration and reduce impact, we introduce a feed-forward backlash bias voltage $u_b$ that applies opposite preload torque to the two motors for backlash compensation, and a high-frequency dithering term $u_d$ that overcomes static friction.}
    \label{fig:block-diagram}
    \vspace{-15pt}
\end{figure}

Low-cost brushed servo motors introduce two control challenges: gear backlash that causes positioning lag and oscillations during direction reversal, and high static friction that prevents the controller from actuating small corrections, resulting in persistent steady-state error. To mitigate these limitations, we developed a dual-motor cooperative control method. Our method integrates dual-motor counter-drive backlash control and static friction compensation. The system's control block diagram is presented in~\cref{fig:block-diagram}.

\textbf{Dual-Motor Counter-Drive Backlash Elimination:} Gear backlash introduces positioning lag during direction reversal, which under high-gain control can excite oscillations. We directly couple the output shafts of two motors and apply opposing bias torques so that each motor engages opposite gear faces, keeping the teeth in constant contact. A feed-forward bias voltage $u_b$ is added to the control loop to maintain this preload. Accordingly, the drive voltages of the two motors are formulated as follows:
\begin{equation}
\left\{
\begin{aligned}
u_1 &= u_o + u_b,\\
u_2 &= u_o - u_b.
\end{aligned}
\right.
\label{eq:counter-drive}
\end{equation}

\textbf{Stiction Compensation through Motor Dithering:} The friction model, combining both Coulomb and viscous elements, is described by the following equation~\cite{olssonFrictionModelsFriction1998}:
\begin{equation}
\tau_f=
\begin{cases}
\tau_s\operatorname{sgn}(\dot q)+\tau_v \dot q, & \dot q\neq 0,\\
\tau_e, & \dot q=0\ \text{and}\ |\tau_e|<\tau_s,\\
\tau_s\operatorname{sgn}(\tau_e), & \text{otherwise}.
\end{cases}
\label{eq:friction-model}
\end{equation}
$\tau_s$ denotes the maximum Coulomb friction; $\tau_v$ represents the viscous friction coefficient; $\tau_e$ signifies the external torque, and $q$ indicates the angle. Initial rotation necessitates overcoming the static threshold $\tau_s$. Low-cost motors exhibit higher $\tau_s$, resulting in the controller's incapacity to generate torque for motor actuation when position error remains minimal, thereby inducing persistent steady-state error.

A common approach to mitigate this issue is to introduce an integral controller. However, the accumulation of the integral term requires time, and due to constraints such as communication time, our frequency of the PID control cycle is relatively low (66 Hz). Therefore, we introduce a feed-forward term to the output:
\begin{equation}
    u_d = (-1)^{\lfloor t/T \rfloor}u_s
    \label{eq:dithering-model}
\end{equation}
where $T$ represents the cycle time of the PID loop, and $u_s$ is the feed-forward term set to maintain the motor in a near-threshold state.

\textbf{Trapezoidal Profile:} Mechanical structures inherently exhibit elasticity, whereby excessive acceleration induces undesirable oscillatory behavior. To address this, a trapezoidal profile was introduced for all joints, including the robotic arm joints, the lifting slide, and the BLDC on the mobile base.
A trapezoidal profile is used as a filter to transform the teleoperation or learning-based position reference $q_r$ into a $C^1$ trajectory command $q_s(t)$, thereby reducing oscillations caused by nonlinear effects under torque saturation. Let the current command at replanning time be $q_{s,0}$, define $\Delta q=q_r-q_{s,0}$ and $s=\operatorname{sgn}(\Delta q)$. The transition from $q_{s,0}$ to $q_r$ is decomposed into acceleration, constant-velocity, and deceleration phases. Given acceleration limit $a_r$ and cruise-speed limit $v_r$, the phase durations are
\begin{equation}
t_a=\frac{v_r}{a_r},\qquad
t_u=\frac{|\Delta q|}{v_r}-\frac{v_r}{a_r},\qquad
t_d=\frac{v_r}{a_r}.
\end{equation}
The smoothed position command is then given by
\begin{equation}
q_s(t)=q_{s,0}+s\,\phi(t),
\end{equation}
\begin{equation}
\phi(t)=
\begin{cases}
\dfrac{1}{2}a_r t^2, & 0\le t<t_a,\\
v_r\!\left(t-\dfrac{1}{2}t_a\right), & t_a\le t<t_a+t_u,\\
|\Delta q| - \dfrac{1}{2}a_r (t_a+t_u+t_d-t)^2, & t_a+t_u\le t<t_a+t_u+t_d,\\
\end{cases}
\end{equation}
For $t\ge t_a+t_u+t_d$, the command is held at $q_s(t)=q_r$. This procedure is updated online whenever a new reference $q_r$ arrives, and phase durations are recomputed from the current command $q_s$. This approach minimizes vibrations and impacts, yielding a smooth motion trajectory.

\section{\tele Teleoperation}

\begin{figure}[!htbp]
    \centering
    \includegraphics[width=0.7\linewidth]{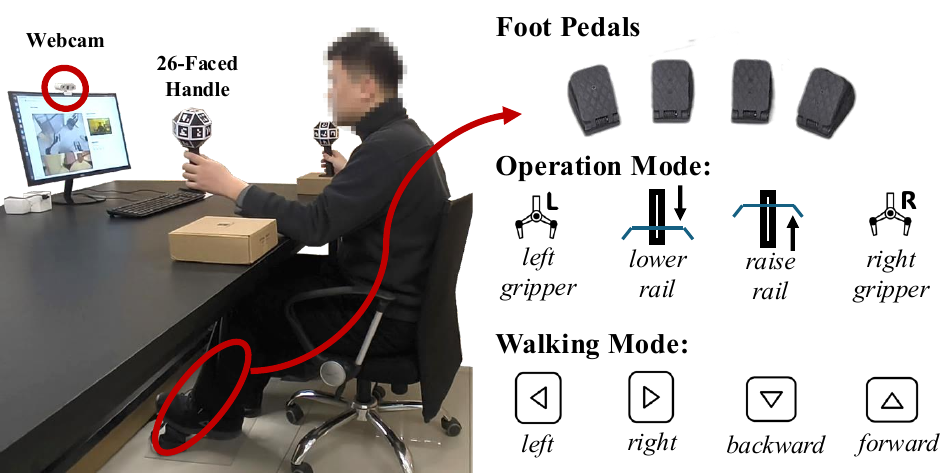}
    \caption{\textbf{\tele Teleoperation Workstation.} By capturing the 6D pose of the handle through a webcam, we can fully remotely teleoperate the robot, with only \$\textit{50}. Foot pedals can switch between two modes, respectively controlling the base's movement and the upper limbs' operation.}
    \label{fig:teleop-workstation}
\end{figure}

We target a dual-arm mobile teleoperation system that is easy to set up, low-cost, and accurate enough for fully remote operation, thereby enabling broader participation in data collection to mitigate data scarcity in embodied AI. We propose \tele, a system comprising two 26-faced marker-based handles, four Hall-effect pedals, a webcam, and an ESP32 microcontroller. As shown in~\cref{fig:teleop-workstation}, the entire workstation costs only \textit{\$50} and requires no head-mounted display, making it suitable for long-duration fully remote teleoperation. The two handles capture the 6-DoF poses of the operator's left and right hands, which are then retargeted to the joint space of the robotic arms via inverse kinematics. The four Hall-effect pedals control the movement of the robot's base, the opening and closing of the left and right grippers, and the large-scale movement of the lifting slider.

\subsection{26-Faced Motion Capture Handle}
\label{sec:26-faced-handle}

\begin{figure}[!htbp]
    \centering
    \subfloat[6-Faced Handle]{\includegraphics[width=0.35\linewidth]{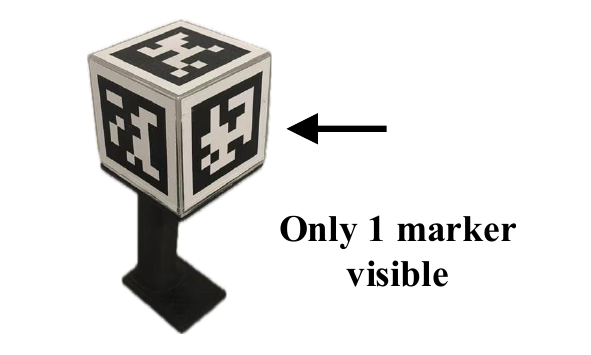}\label{fig:handle_compare-6}}\quad
    \subfloat[26-Faced Handle (Proposed)]{\includegraphics[width=0.35\linewidth]{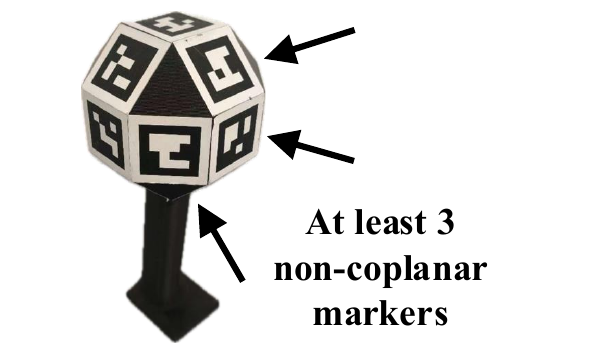}\label{fig:handle_compare-26}}
    \caption{\textbf{Comparison Between 6-Faced and 26-Faced Handles.} The conventional 6-faced handle may expose only one visible marker when a face is nearly parallel to the camera plane, which induces pose ambiguity problem and can produce abrupt pose switching due to multiple valid solutions. In contrast, the proposed 26-faced handle maintains at least three simultaneously visible non-coplanar markers across viewpoints, mitigating ambiguity and preventing discontinuous pose estimation.}
    \label{fig:handle_compare}
\end{figure}

\begin{figure}[!htbp]
    \centering
    \subfloat[2D image-plane projection]{\includegraphics[width=0.35\linewidth]{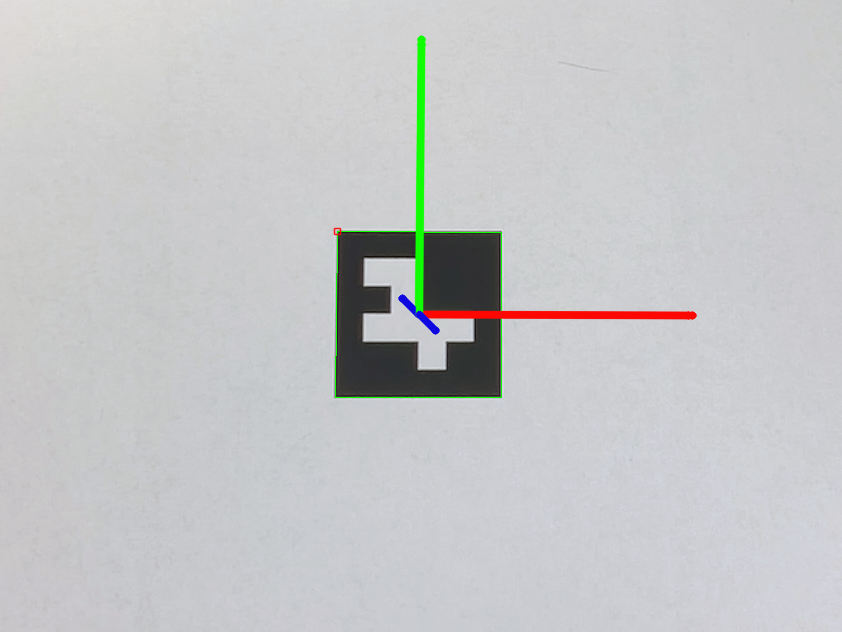}\label{fig:ippe-two-sol-2d}}\quad
    \subfloat[3D pose visualization]{\includegraphics[width=0.35\linewidth]{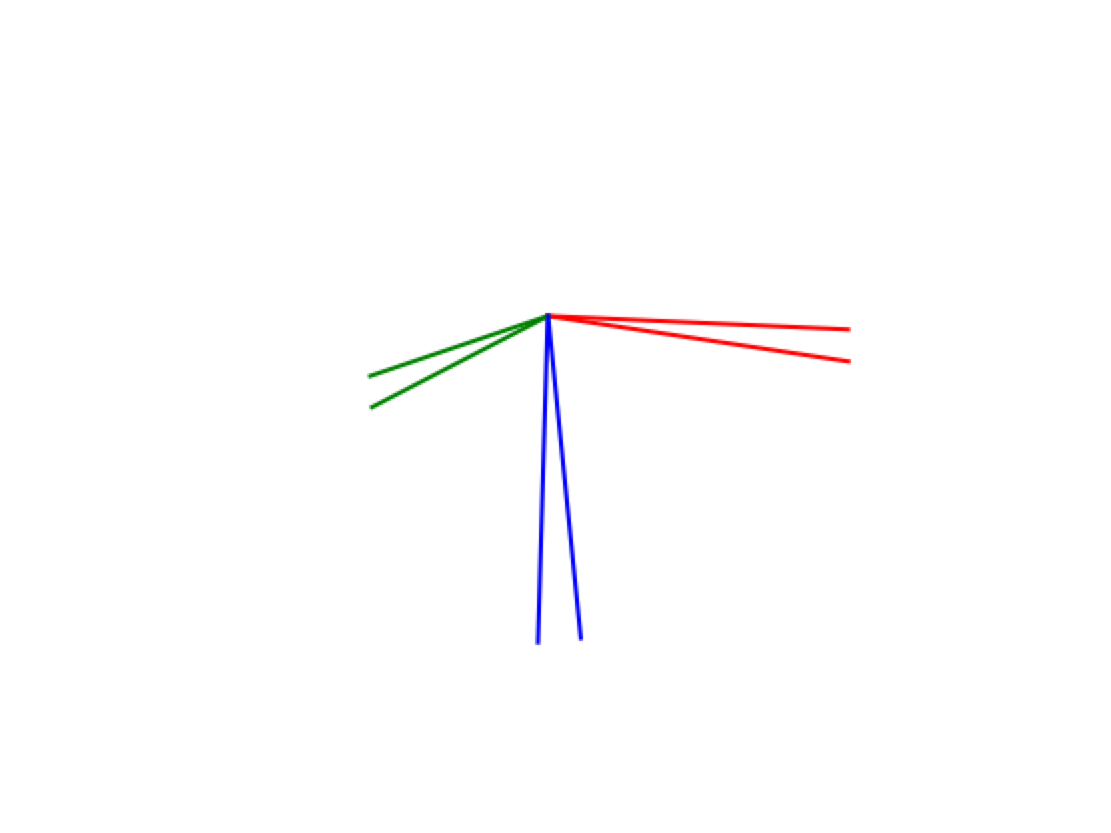}\label{fig:ippe-two-sol-3d}}
    \caption{\textbf{Visualization of pose ambiguity in planar PnP.} The two candidate poses are obtained by IPPE~\cite{collinsInfinitesimalPlaneBasedPose2014}. In typical cases, the solution with larger reprojection error can be rejected. However, under near-affine projection, the two candidates can yield comparable reprojection errors, causing ambiguous pose selection and rotation spikes. Introducing multiple non-coplanar markers increases geometric constraints and mitigates this ambiguity.}
    \label{fig:ippe-two-solution}
\end{figure}

To command robot actions, we estimate the desired Cartesian pose of each arm end-effector from a uniquely designed 26-faced handle~\cref{fig:handle_compare-26}. The handle pose estimation is formulated as a Perspective-$n$-Point (PnP) problem, i.e., estimating the rigid transformation ${}^{h}_{c}\mathbf{T}\in SE(3)$ between the camera frame and the handle frame. Specifically, multiple AprilTags~\cite{olsonAprilTagRobustFlexible2011} are attached to the handle at predefined locations. We detect 2D tag corners in the image, $\mathbf{u}_i=[u_i,v_i,1]^\top\in\mathbb{P}^2$, and associate them with known 3D reference points on the handle, $\tilde{\mathbf{P}}_i=[x_i,y_i,z_i,1]^\top\in\mathbb{P}^3$.

We then solve for the pose by minimizing the reprojection error using the Levenberg--Marquardt algorithm:
\begin{equation}
{}^{h}_{c}\mathbf{T}^{*}=\arg\min_{\mathbf{T}\in SE(3)}\sum_{i=1}^{n}\left\lVert\mathbf{u}_i-\pi(\mathbf{K},\mathbf{T},\tilde{\mathbf{P}}_i)\right\rVert_{\Sigma}^{2},
\end{equation}
where the projection function and camera intrinsic matrix are defined as follows:
\begin{equation}
\pi(\mathbf{K},\mathbf{T},\tilde{\mathbf{P}}_i)=\frac{1}{Z_{c,i}}\mathbf{K}[\mathbf{I}\mid\mathbf{0}]\mathbf{T}\tilde{\mathbf{P}}_i,
\end{equation}
\begin{equation}
\mathbf{K}=\begin{bmatrix}
 f_x & 0 & c_x \\
 0 & f_y & c_y \\
 0 & 0 & 1
\end{bmatrix}.
\end{equation}

A common alternative is a 6-faced cube handle~\cref{fig:handle_compare-6}. However, PnP with coplanar points is known to suffer from pose ambiguity~\cite{schweighoferRobustPoseEstimation2006}, especially when visible tags are nearly parallel to the image plane. This phenomenon is visualized in~\cref{fig:ippe-two-solution}, where two candidate poses obtained by IPPE~\cite{collinsInfinitesimalPlaneBasedPose2014} are shown. Under near-affine projection, the two solutions can produce comparable reprojection errors, making it difficult to reject one by reprojection error alone, which leads to rotation spikes and degraded tracking accuracy.

To address this issue, we propose a 26-faced polyhedral marker layout by inserting additional faces between neighboring cube faces while keeping the handle compact. The 26 faces correspond to the complete 26-connected neighborhood of a cube, maximizing non-coplanar marker orientations within a minimal volume. This design increases non-coplanar visual constraints so that, from arbitrary viewpoints, the camera can typically observe at least three non-coplanar tags simultaneously. Consequently, the proposed handle substantially mitigates pose ambiguity and provides more stable, higher-precision pose estimation.

\subsection{Kinematics}

The estimated end-effector pose in $SE(3)$ is retargeted to joint space by solving an inverse-kinematics (IK) problem. At each step, this is achieved by solving the optimization problem:
\begin{equation}
\mathbf{q}^{*}=\arg\min_{\mathbf{q}}\left\|\operatorname{Log}\!\left(\mathbf{T}(\mathbf{q})^{-1}\mathbf{T}_{d,t}\right)^{\vee}\right\|_{2}^{2},
\end{equation}
where $\mathbf{T}(\mathbf{q})\in SE(3)$ denotes the forward kinematics, $\operatorname{Log}(\cdot)$ is the matrix logarithm on $SE(3)$, and $(\cdot)^{\vee}$ maps elements from $\mathfrak{se}(3)$ to their vector form. The desired end-effector pose is defined by the handle-referenced transformation chain,
\begin{equation}
\mathbf{T}_{d,t}={}^{b}_{e}\mathbf{T}_{0}\,{}^{h}_{c}\mathbf{T}_{0}\,\left({}^{h}_{c}\mathbf{T}_{t}\right)^{-1},
\end{equation}
where $t=0$ is the initialization time at which the handle frame and the robot tool frame are aligned.

\subsection{Web-based Remote Teleoperation Interface}

Four Hall-effect pedals capture operator foot inputs and support two modes, switchable via keyboard shortcuts (see~\cref{fig:teleop-workstation}): a walking mode for base translation and rotation, and an operation mode for gripper open/close and lift control. During base movement, the gripper position can be locked to ensure stable object gripping. The entire teleoperation client runs in a standard web browser. On the client side, 26-faced-handle detection and pose estimation are executed locally via WebAssembly and OpenCV.js, while pedal data are relayed through WebSerial by an ESP32 microcontroller. On the robot side, three camera feeds are streamed to the browser over WebRTC, and the handle 6-DoF pose together with pedal commands are returned through a WebRTC DataChannel.

\section{Experiment}

We evaluate the proposed system across five dimensions: control precision, teleoperation interface, remote task execution, teleoperation efficiency, and imitation learning. Our evaluation is guided by the following research questions:

\begin{itemize}
\item \textbf{RQ1: Control Precision ---} To what extent does the dual-motor backlash elimination with dithering enhance the end-effector accuracy and repeatability of low-cost hardware?
\item \textbf{RQ2: Interface Stability and Resolution ---} Does the proposed 26-face handle improve tracking stability and spatial resolution relative to a conventional 6-face handle?
\item \textbf{RQ3: Remote Capability ---} Under fully remote operation, can our system complete fine-grained, complex, and long-horizon household tasks requiring whole-body mobility without on-site assistance?
\item \textbf{RQ4: Teleoperation Efficiency ---} Compared with mainstream teleoperation methods, can \tele achieve comparable or better success rate and operational efficiency at a substantially lower cost?
\item \textbf{RQ5: Learning Effectiveness ---} Are the demonstrations collected via \tele sufficient to train high-quality policies for end-to-end autonomous task execution in tasks requiring bimanual coordination and whole-body control?
\end{itemize}

\begin{figure}[!htbp]
    \centering
    \subfloat[Stair wave response, w/ and w/o dithering enabled]{
        \includegraphics[width=0.3\linewidth]{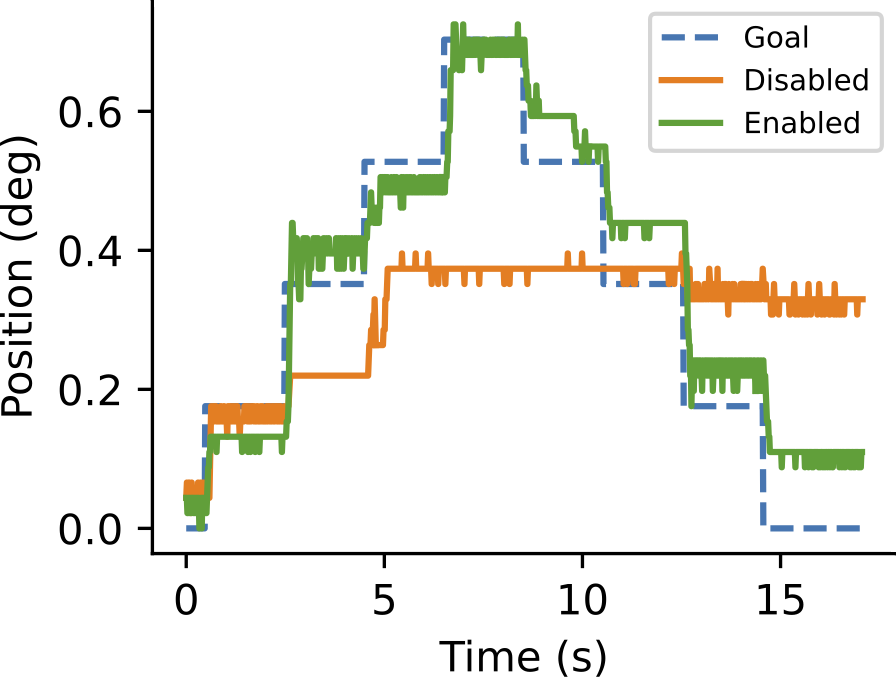}
        \label{fig:stiction-stairwave}
    }
    \quad
    \subfloat[Square wave response, w/ and w/o counter-drive enabled]{
        \includegraphics[width=0.3\linewidth]{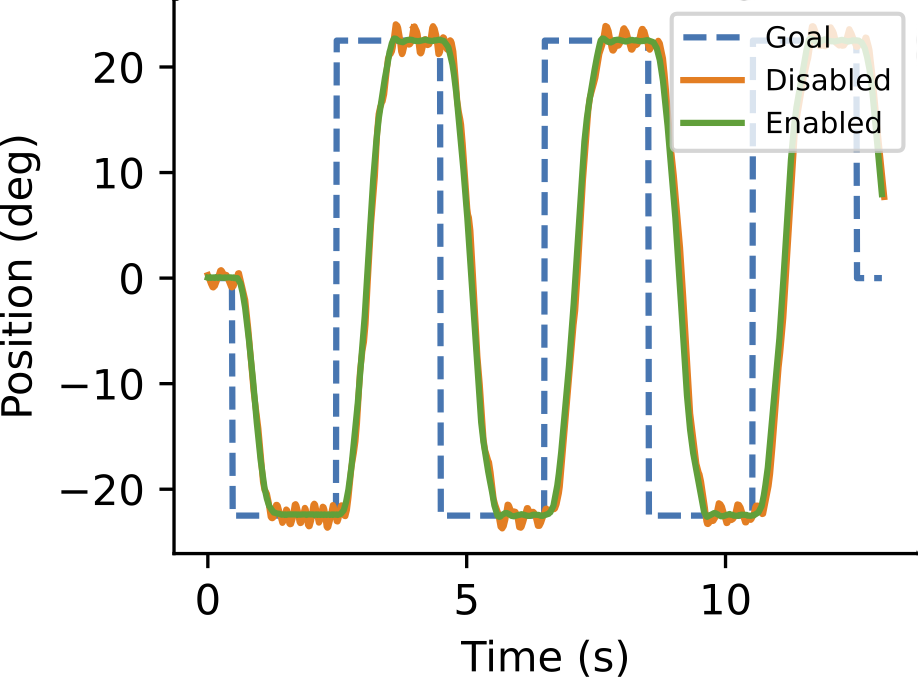}
        \label{fig:backlash-squarewave}
    }
    \caption{\textbf{System Responses with Different Control Strategies.} Tracking responses under motor dithering and counter-drive backlash compensation are compared, highlighting improved micro-step tracking and reduced oscillations.}
    \label{fig:response-comparison}
    \vspace{-10pt}
\end{figure}

\begin{figure}[!htbp]
    \centering
    \subfloat[Gripper contacts a dial indicator]{\includegraphics[width=0.35\linewidth]{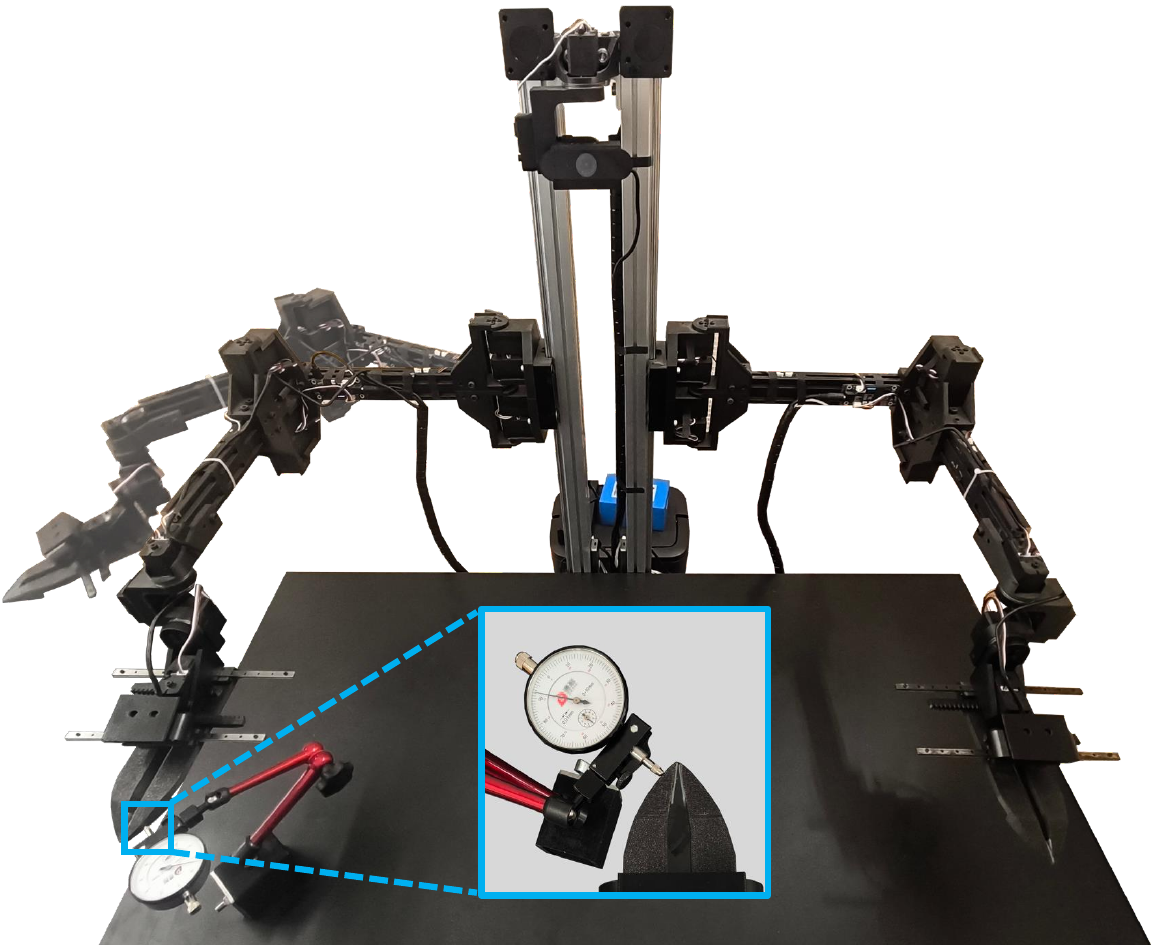}\label{fig:repeatability-setup}}\quad
    \subfloat[Repeatability error histogram]{\includegraphics[width=0.35\linewidth]{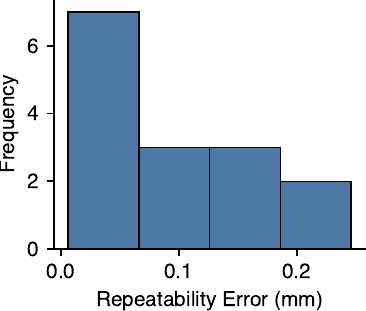}\label{fig:repeatability-histogram}}
    \caption{\textbf{Repeatability evaluation.} A dial indicator measures end-effector positioning repeatability, and deviations from the mean are reported in the histogram.}
    \label{fig:repeatability}
\end{figure}

\subsection{Comparison of Control Strategies}

We conducted experiments to evaluate micromotion tracking capabilities, employing a staircase trajectory with incremental steps of 0.175 degrees, twice the motor's minimum resolution. The results~(\cref{fig:stiction-stairwave}) showed that with motor dithering enabled, the system successfully tracked incremental target adjustments, whereas the control group with motor dithering disabled failed to follow the target. In addition, we compared the performance of enabling and disabling the counter-drive backlash elimination module. The results~(\cref{fig:backlash-squarewave}) illustrate the tracking of a square wave target by recording the actual positions reported by the position sensor installed on the motor. When the module is disabled, the joint exhibits oscillations around the target point due to the presence of backlash. This method successfully suppresses oscillations, which enhances positioning accuracy.

\subsection{Repeatability Evaluation}

End-effector positioning repeatability was evaluated using a dial indicator setup as illustrated in \cref{fig:repeatability-setup}. To ensure measurement integrity, the robot was secured to the test platform, eliminating base movement. The gripper performed 15 unidirectional approach cycles toward the indicator; the readings yielded a mean of $\bar{x}=5.79$\,mm with a standard deviation of $\sigma=0.12$\,mm. Following the $\pm 3\sigma$ metric, the repeatability is reported as $0.72$ mm, with the error distribution visualized in \cref{fig:repeatability-histogram}. These results confirm sub-millimeter repeatability, supporting the system's capability for high-fidelity motion and data collection in household environments.

\FloatBarrier
\subsection{Tracking Accuracy of 6-Faced vs. 26-Faced Handles}

\begin{figure}[!htbp]
    \centering
    \includegraphics[width=0.6\linewidth]{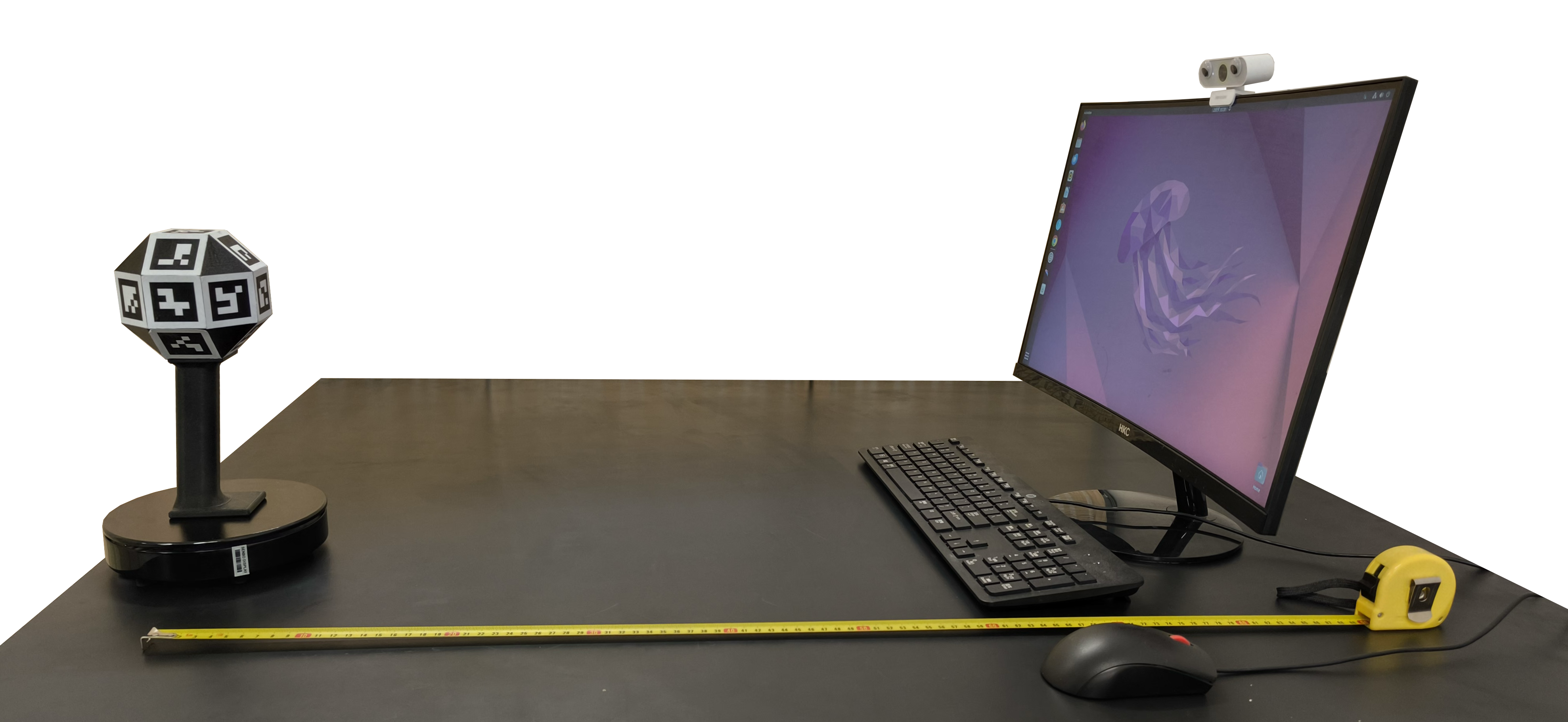}
    \caption{\textbf{Experimental Setup for Comparing Various Handles.} The handles are mounted on a turntable; we rotate it for three full turns and record position and rotation errors.}
    \label{fig:face-handle-exp-setup}
\end{figure}

\begin{figure}[!htbp]
    \centering
    \includegraphics[width=0.6\linewidth]{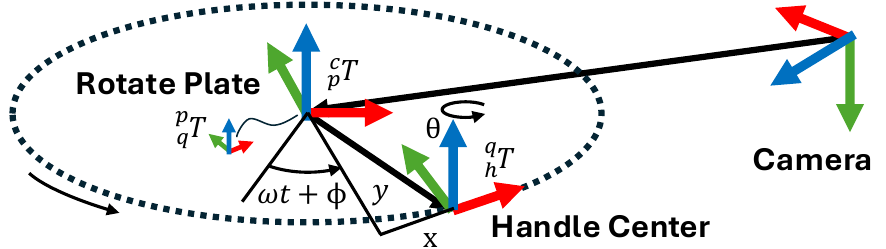}
    \caption{\textbf{Model of the Experimental Setup for Handles.} We align the handle as close as possible to the turntable center; installation errors (e.g., eccentricity) are modeled and removed to enable fair comparison.}
    \label{fig:face-handle-exp-setup-explain}
\end{figure}

To quantitatively assess the errors associated with the 6-faced and the 26-faced polyhedron, we positioned our tag on a rotating platform as shown in~\cref{fig:face-handle-exp-setup} and performed three complete rotations. Both positional and rotational errors were recorded. A camera was placed 800 mm away from the rotating platform to capture images at a resolution of 1920x1080. The actual transformation matrix $\hat{_h^cT}$ was obtained by solving this Perspective-n-Point problem. Although the handle was installed as centrally as possible on the rotating platform, some installation offsets persisted. To mitigate these errors in the experimental results, we modeled the handle's rotation as occurring around a circular path, as illustrated at~\cref{fig:face-handle-exp-setup-explain}.

\begin{figure}[!htbp]
    \centering
    \subfloat[Translation Error]{
        \includegraphics[width=0.6\linewidth]{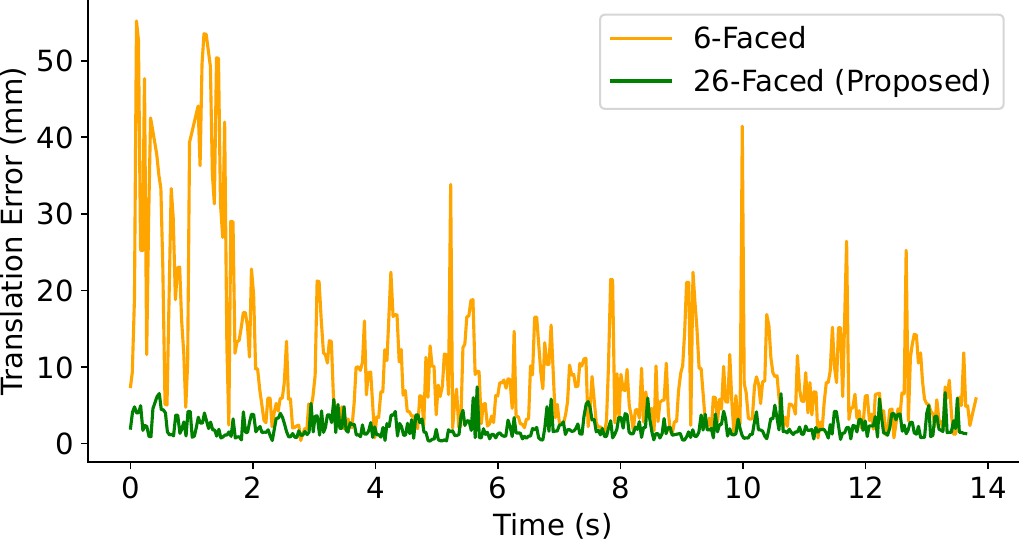}
        \label{fig:robocu-pos-err}
    }
    \\[3pt]
    \subfloat[Rotation Error]{
        \includegraphics[width=0.6\linewidth]{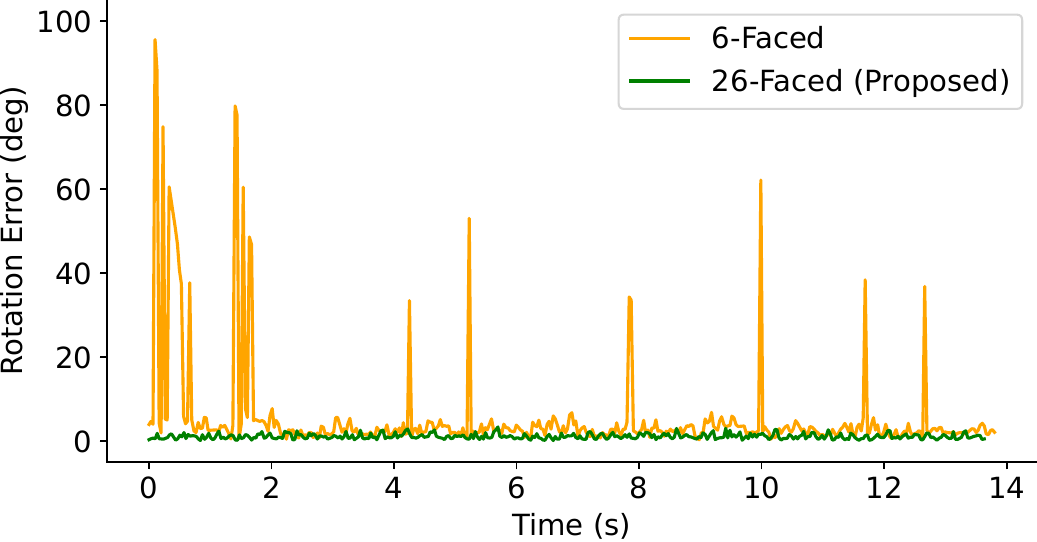}
        \label{fig:robocu-rot-err}
    }
    \caption{\textbf{Tracking Accuracy of Different Handles.} Our proposed 26-faced handle reduces error by 80\% compared to the traditional design and avoids the rotation-error spikes observed with the 6-faced handle.}
    \label{fig:err-comparison}
\end{figure}

The ideal transformation from handle to camera is modeled as ${_h^cT} = {_p^cT} \cdot {_q^pT}(t) \cdot {_h^qT}$, where ${_p^cT} = \exp[\mathbf{v}]_\times$ is the fixed location of rotation plate, ${_q^pT}(t) = R_z(\omega t + \phi)$ describes rotation over time, and ${_h^qT} = R_z(\theta) + [x\ y\ 0]^T$ accounts for mounting offsets in the x-y plane. The installation offsets parameters $x, y, \theta, \omega, \phi, \mathbf{v}$ are estimated by minimizing the sum of positional and rotational differences~\cite{hartleyRotationAveraging2013} to the ideal transformation, $\sum_t \left[ \| P_t - \hat P_t \|_2 + d_\angle(R_t, \hat R_t) \right]$, using Adam optimization over 2000 steps, where $P_t$ and $R_t$ are the translation and rotation components of ${_h^cT}_t$. The resulting differences for both the 6-faced and 26-faced polyhedra are shown in~\cref{fig:err-comparison}, with average values summarized in~\cref{tab:errors}.

\begin{table}[!htbp]
    \centering
    \caption{Tracking accuracy comparison between 6-faced and 26-faced handles.}
    \label{tab:errors}
    \begin{tabular}{@{}lcc@{}}
        \toprule
        \textbf{Handle Type} & \textbf{Avg. Rotation Err ($^\circ$)} & \textbf{Avg. Translation Err (mm)} \\
        \midrule
        6-Faced & 5.391 & 9.9 \\
        \textbf{26-Faced (Ours)} & \textbf{1.094} \down 80\% & \textbf{2.1} \down 79\% \\
        \bottomrule
    \end{tabular}
\end{table}

As shown in~\cref{fig:err-comparison} and~\cref{tab:errors}, the 26-faced handle reduces the average rotation error from $5.391^\circ$ to $1.094^\circ$ (80\% reduction) and the average translation error from 9.9\,mm to 2.1\,mm (79\% reduction). Notably, the 6-faced handle exhibits periodic error spikes that coincide with viewpoints where visible markers become nearly coplanar, whereas the 26-faced handle produces smooth, spike-free tracking throughout all three rotations. This is because the 26-faced design ensures that at least three non-coplanar tags are visible from any viewpoint, effectively eliminating the pose ambiguity inherent in planar PnP.

\FloatBarrier
\subsection{Spatial Resolution Evaluation}

\begin{figure}[!htbp]
    \centering
    \includegraphics[width=0.6\linewidth]{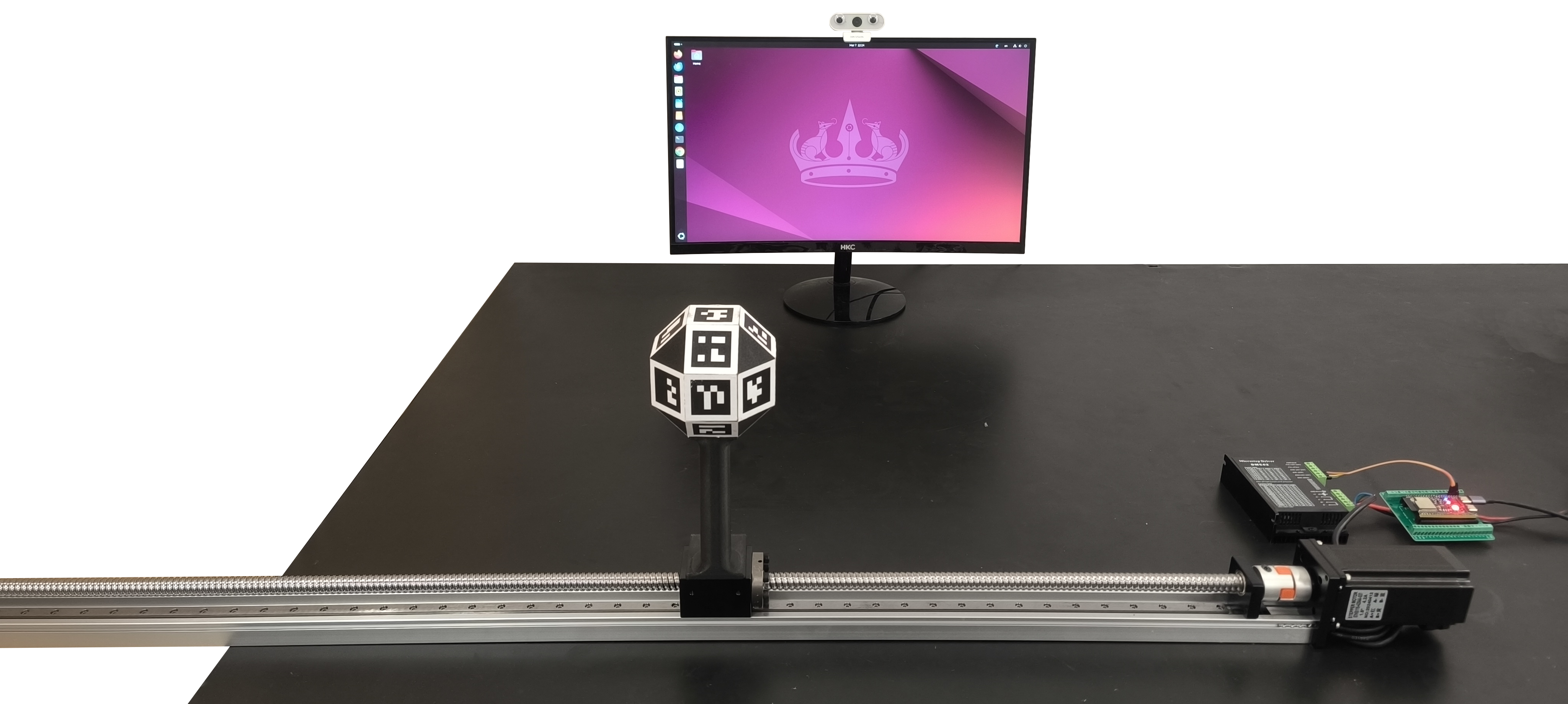}
    \caption{\textbf{Resolution Experiment Setup.} A ball-screw linear guide provides Ground Truth displacement. We synchronously record the Ground Truth position and the position estimated from the 26-faced marker handle.}
    \label{fig:resolution-exp-setup}
\end{figure}

To verify precision for fine teleoperation, we conducted a spatial-resolution experiment. Following the rotational-error setup, the camera was placed about 80 cm from the 26-faced handle (typical operating distance) as shown in~\cref{fig:resolution-exp-setup}. A ball-screw linear guide was rigidly attached to the handle to provide Ground Truth displacement. Images were captured at 1920\(\times\)1080. The guide was connected to the host computer via a stepper driver and an ESP32 controller. During collection, the computer recorded RoboPilot-estimated handle pose ${_t^cT}$ and true guide displacement $d$, and we report estimated displacement $\hat{d}$ versus $d$. The guide was commanded to move 1 mm every 2 seconds, yielding a staircase trajectory.

\begin{figure}[!htbp]
    \centering
    \includegraphics[width=0.7\linewidth]{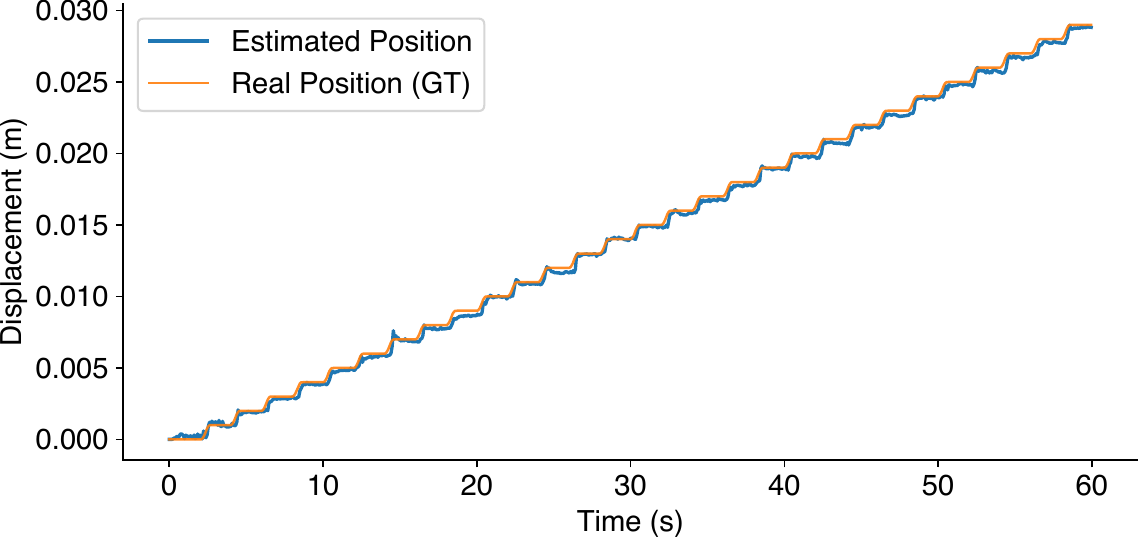}
    \caption{\textbf{Spatial Resolution Evaluation.} With corner refinement enabled, the 26-faced handle achieves approximately 1\,mm spatial resolution. The estimated position (blue) closely tracks the ground-truth staircase displacement (orange) with low noise and small cumulative error.}
    \label{fig:resolution-with-corner-refine}
\end{figure}

As shown in~\cref{fig:resolution-with-corner-refine}, with corner refinement enabled the 26-faced handle achieves approximately 1\,mm spatial resolution: the estimated displacement closely follows the ground-truth staircase trajectory with low noise and small cumulative error. This confirms that the proposed handle provides sufficient precision for fine-grained teleoperation and high-quality offline data collection.

\FloatBarrier
\subsection{Comparison of Teleoperation Methods}

\begin{figure}[!htbp]
    \centering
    \includegraphics[width=0.7\linewidth]{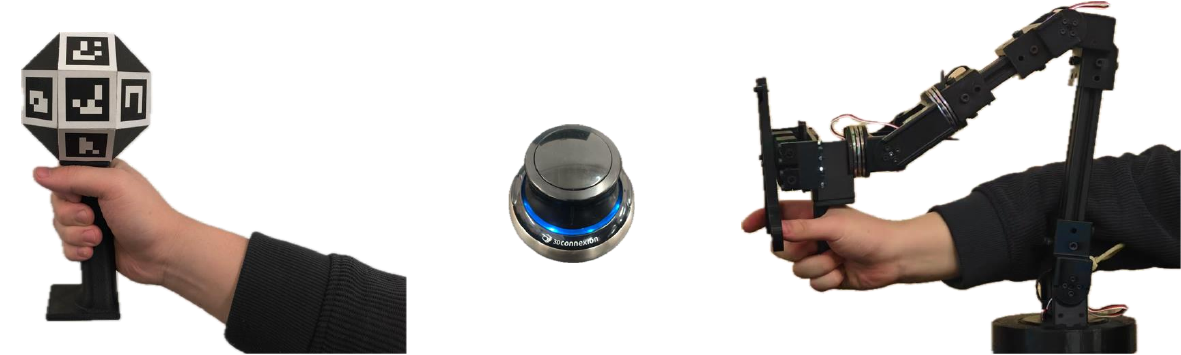}
    \caption{\textbf{Comparison of Teleoperation Methods.} We compare three teleoperation methods: our proposed 26-faced handle, a SpaceMouse, and a leader--follower robotic arm.}
    \label{fig:teleop-compare}
\end{figure}

\begin{figure}[!htbp]
    \centering
    \includegraphics[width=0.7\linewidth]{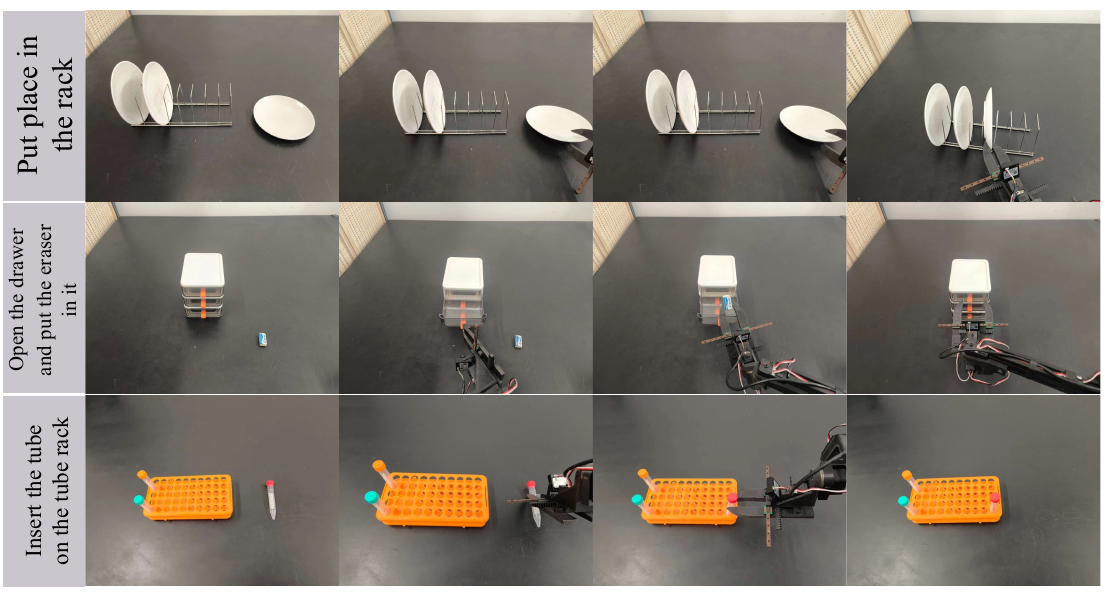}
    \caption{\textbf{Teleoperation Tasks Demonstration.} Three challenging fine-grained tasks are used to evaluate teleoperation methods and require precise object manipulation: inserting a plate into a rack, grasping a 1~cm$^2$ size drawer handle, and inserting a test tube into a space-constrained target.}
    \label{fig:teleop-tasks}
\end{figure}

\tele was compared with two commonly used approaches: Leader-Follower~\cite{fuMobileALOHALearning2024} and SpaceMouse~\cite{luoPreciseDexterousRobotic2024}. For the leader-follower method, an early design prototype robotic arm shown in~\cref{fig:teleop-compare} was adopted.
The SpaceMouse was configured to track the end-effector poses' position and rotation changes relative to the robot's Base Frame. To ensure fair comparison, teleoperators were required to operate while seated at a computer and observe remote images. Each participant had one warm-up opportunity, followed by the formal experiment, continuing until five sets of successful data were collected. We selected 3 tasks and recorded the success rate and the average time of successful attempts. The task examples and full results are shown in \cref{fig:teleop-tasks} and \cref{tab:task_comparison}.

\begin{table}[!htbp]
    \small
    \caption{Task Performance Comparison across Different Methods.\\ \textbf{SR:} success rate. \textbf{Time:} average success time (s).}
    \label{tab:task_comparison}
    \centering
    \begin{tabular}{@{}ccc|cc|cc@{}}
        \toprule
        \multirow{2}{*}{\textbf{Task Name}\vspace{-2mm}} & \multicolumn{2}{c|}{\textbf{\tele}} & \multicolumn{2}{c|}{\textbf{SpaceMouse}} & \multicolumn{2}{c}{\textbf{Leader-Follower}} \\
        \cmidrule(lr){2-3} \cmidrule(lr){4-5} \cmidrule(lr){6-7}
        & \textbf{SR} & \textbf{Time} & \textbf{SR} & \textbf{Time} & \textbf{SR} & \textbf{Time} \\
        \midrule
        Put plate in the rack & 100\% & \textbf{36.62} & 44.4\% & 47.31 & 100\% & 57.35 \\
        Open drawer and put eraser in & 100\% & \textbf{72.60} & 100\% & 83.95 & 100\% & 82.69 \\
        Insert the tube on the rack & 100\% & \textbf{26.44} & 100\% & 44.87 & 100\% & 55.94 \\
        \midrule
        Average & 100\% & \textbf{45.22} & 81.5\% & 58.71 & 100\% & 65.33 \\
        \midrule
        Hardware Cost & \multicolumn{2}{c|}{\$50} & \multicolumn{2}{c|}{\$220} & \multicolumn{2}{c}{\$260} \\
        \bottomrule
    \end{tabular}
\end{table}

As shown in~\cref{tab:task_comparison}, \tele reduces average completion time by 30\% over the baselines while achieving 100\% success rate across all tasks. The SpaceMouse suffers from coordinate-frame confusion when operators switch between the head and wrist cameras, increasing operation time; its incremental control also introduces dead zones, lowering the success rate on Task~1 to 44.4\%. The leader-follower arm frequently encounters wrist singularities due to its roll-pitch-roll configuration, requiring manual intervention to recover; replacing joints with encoder-only modules or brushless motors can reduce resistance but substantially increases cost and still suffers from workspace mismatches. In contrast, \tele avoids these issues through direct Cartesian-space handle tracking.

\FloatBarrier
\subsection{Very Long-horizon Remote Teleoperation}

\begin{figure}[!htbp]
    \centering
    \includegraphics[width=1\linewidth]{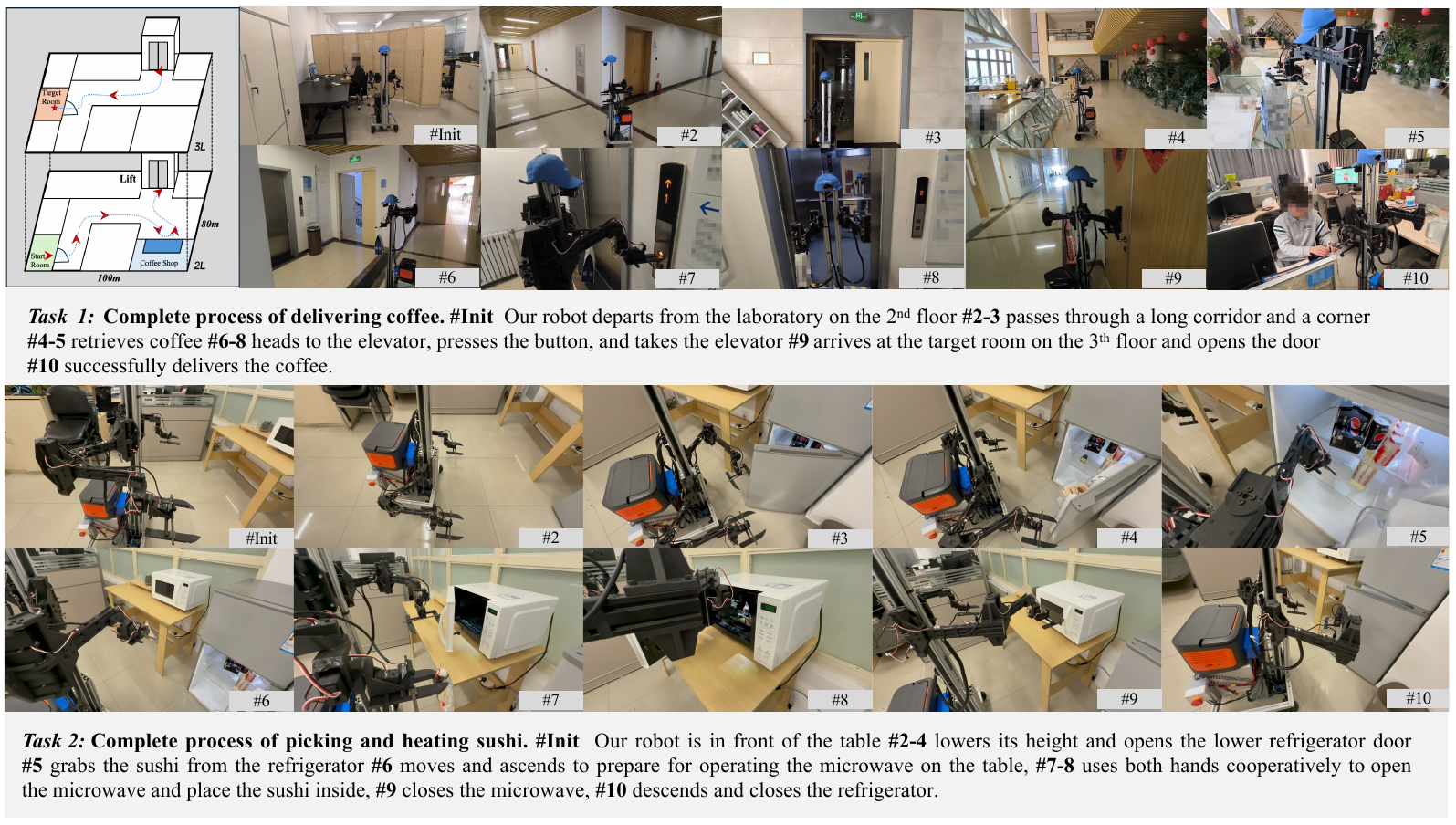}
    \caption{\textbf{Very Long-horizon Remote Teleoperation.} We used \tele to control \rob and demonstrated two specially designed complex tasks. \textit{Task 1} involves a very long sequence of operations with a total movement distance exceeding 200m, requiring remote tele-communication, agile movement, and precise environmental interaction. \textit{Task 2} specifically showcases \rob's lifting and lowering capabilities, enabling it to touch the ground and complete more complex tasks in everyday life. We also demonstrated \rob's bimanual coordination by opening a microwave and placing sushi inside.}
    \label{fig:complex_tele}
\end{figure}

To probe the operational limits of the system, we designed two long-horizon tasks that require sustained remote teleoperation, whole-body mobility, and precise environmental interaction (\cref{fig:complex_tele}). \textit{\textbf{Task 1}} involves a complete process of delivering coffee, with a total movement distance exceeding 200\,m, requiring long-term teleoperation, full-body mobility, and precise environmental interactions. \textit{\textbf{Task 2}} involves picking and heating food, requiring a large operational space (e.g., opening a microwave) and vertical mobility (the sushi is located in the lower part of a refrigerator, and the target location is on an upper-level table). Since Mobile Aloha lacks vertical movement capabilities, it is unable to complete Task 2. Both tasks were completed fully remotely via the web-based interface over a cellular network, with no on-site assistance. Videos are available on the project website.

\FloatBarrier
\subsection{Autonomous Task Execution via Imitation Learning}

\begin{figure}[!htbp]
    \centering
    \includegraphics[width=1\linewidth]{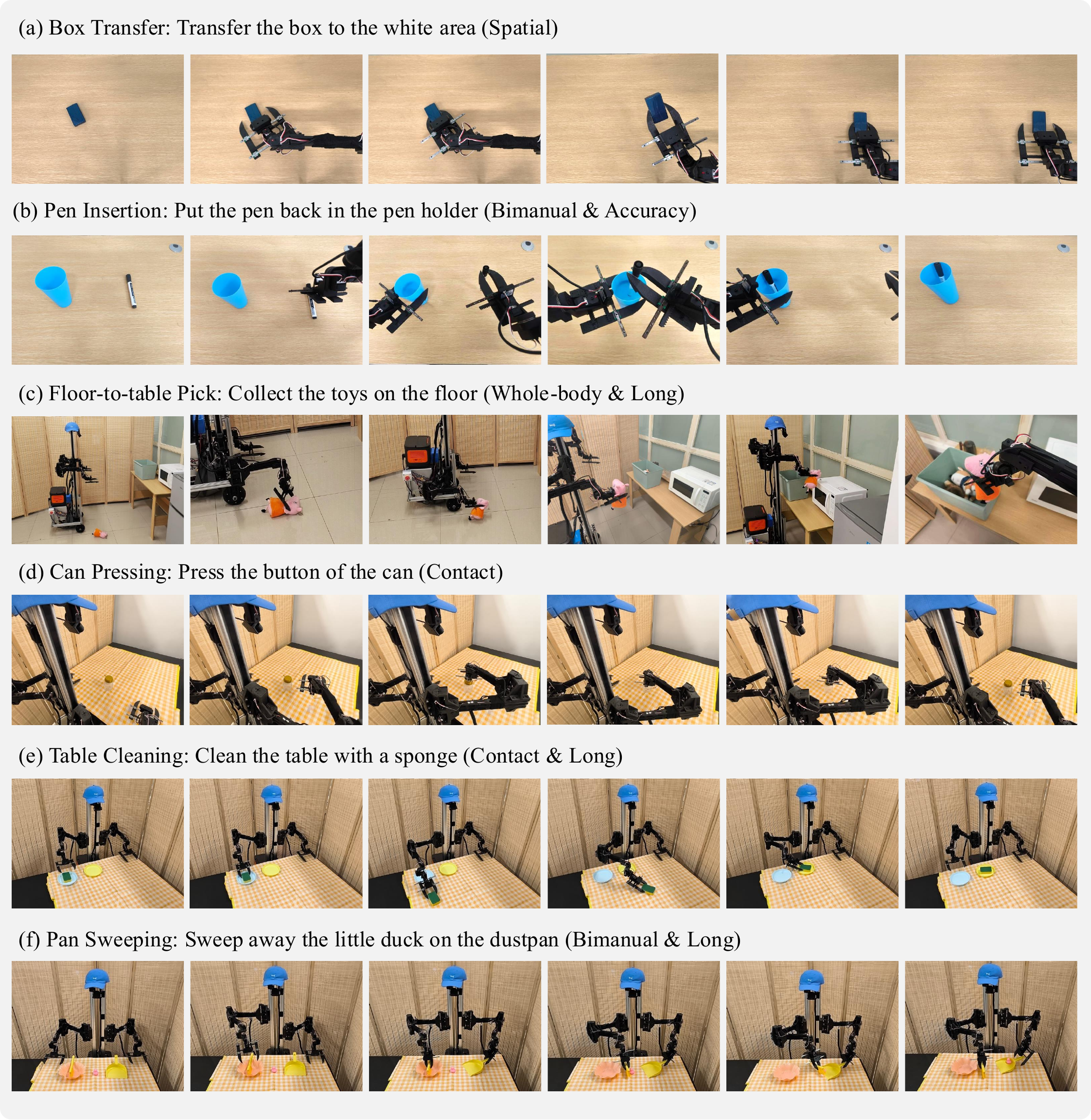}
    \caption{\textbf{Visualization of Imitation Learning Tasks.} The tasks consist of multiple sub-task stages and require changes in height or bimanual coordination.}
    \label{fig:learning-tasks}
\end{figure}

To verify that \rob and \tele can collect demonstrations of sufficient quality for training autonomous policies, we designed six tasks covering bimanual manipulation, whole-body control, contact-rich interaction, and long-horizon operations (illustrated in~\cref{fig:learning-tasks}):
(a) \textbf{Box Transfer}: The robot is required to grasp a small blue box on the table and place it at a designated location with high success.
(b) \textbf{Pen Insertion}: A bimanual coordination task where the right arm picks up a pen and the left arm picks up a cup. The robot must insert the pen into the cup and return the cup to the table without tipping, testing its precise alignment and dual-arm synchronization.
(c) \textbf{Floor-to-table Pick}: This task simulates a common "in-the-wild" scenario where an object is dropped on the floor. The robot must lower its body, retrieve the object, and place it back into a bin on the table, demonstrating its full-body mobility and large-scale vertical workspace coordination.
(d) \textbf{Can Pressing}: The right arm locates and presses a can on the table. This task evaluates the robot's capabilities in contact-rich interaction, requiring sufficient downward force on the object.
(e) \textbf{Table Cleaning}: The robot locates a sponge, wipes the table surface, and returns the sponge to a plate. This involves continuous grasping under surface contact.
(f) \textbf{Pan Sweeping}: The robot coordinates the right arm (holding a broom) and the left arm (holding a dustpan) to sweep toy ducks into the pan. This task challenges accurate tool use and long-horizon bimanual coordination.

Two representative imitation learning algorithms were evaluated: ACT~\cite{fuMobileALOHALearning2024} and $\pi_{0}$~\cite{black$p_0$VisionLanguageActionFlow2024}. Each policy takes as input an 18-dimensional robot state (arm joint positions, gripper state, head camera angle, and base movement command) together with three camera images at $640\times360$ resolution. We collected 50 demonstrations for ``Box Transfer'', ``Can Pressing'', and ``Pen Insertion'', 80 for ``Floor-to-table Pick'', and 200 for ``Table Cleaning'' and ``Pan Sweeping''; more demonstrations were collected for tasks involving longer horizons or higher contact complexity to ensure sufficient coverage of the action distribution.

\begin{table}[!htbp]
    \centering
    \small
    \setlength{\tabcolsep}{4pt}
    \renewcommand{\arraystretch}{1.1}
    \caption{Success rates of different algorithms on real-world dexterous tasks.}
    \label{tab:algo_diversity}
    \begin{tabular}{cclc}
        \toprule
        \textbf{Model} & \textbf{Task} & \textbf{Step} & \textbf{SR} \\
        \midrule
        \multirow{8}{*}{ACT} & \multirow{2}{*}{\makecell[c]{Box Transfer\\(Spatial)}} & Reach and grasp box & 10/10 \\
        & & Lift and place box & 10/10 \\
        & \multirow{3}{*}{\makecell[c]{Pen Insertion\\(Bimanual)}} & Approach and pick up pen & 10/10 \\
        & & Grasp and stabilize cup & 6/10 \\
        & & Align and insert pen & 6/10 \\
        & \multirow{3}{*}{\makecell[c]{Floor-to-table Pick\\(Whole-body) \\ (Long)}} & Reach down and grasp toy & 8/10 \\
        & & Navigate and align to table & 7/10 \\
        & & Place toy into box & 7/10 \\
        \midrule
        \multirow{8}{*}{$\pi_0$} & \multirow{2}{*}{\makecell[c]{Can Pressing\\(Contact)}} & Approach and locate can & 10/10 \\
        & & Apply press to button & 6/10 \\
        & \multirow{3}{*}{\makecell[c]{Table Cleaning\\(Contact)}} & Pick up sponge from table & 9/10 \\
        & & Wipe surface with contact & 7/10 \\
        & & Return sponge to start & 7/10 \\
        & \multirow{3}{*}{\makecell[c]{Pan Sweeping\\(Bimanual) \\ (Long)}} & Lift broom with right hand & 8/10 \\
        & & Hold dustpan with left hand & 6/10 \\
        & & Sweep toy duck into pan & 5/10 \\
        \bottomrule
    \end{tabular}
\end{table}

As shown in~\cref{tab:algo_diversity}, each task was evaluated over 10 trials. Simple tabletop tasks such as ``Box Transfer'' achieved 100\% success rate. For ``Pen Insertion'', the lightweight cup was prone to being knocked over, causing a performance drop at the ``Grasp Cup'' step. For the $\pi_{0}$ tasks, ``Can Pressing'' occasionally terminated the press early, possibly because $\pi_{0}$ decides based on the current frame without temporal memory. In ``Table Cleaning'' and ``Pan Sweeping'', the action-chunking mechanism and longer inference time introduced command discontinuities that occasionally misaligned or knocked over the tools. These failures primarily stem from limitations of current learning-based methods rather than the data-collection pipeline. We emphasize that tasks such as ``Floor-to-table Pick'' are infeasible for platforms like Mobile ALOHA that cannot reach the floor, highlighting the value of \rob's vertical mobility.

\begin{figure}[!htbp]
    \centering
    \includegraphics[width=0.7\linewidth]{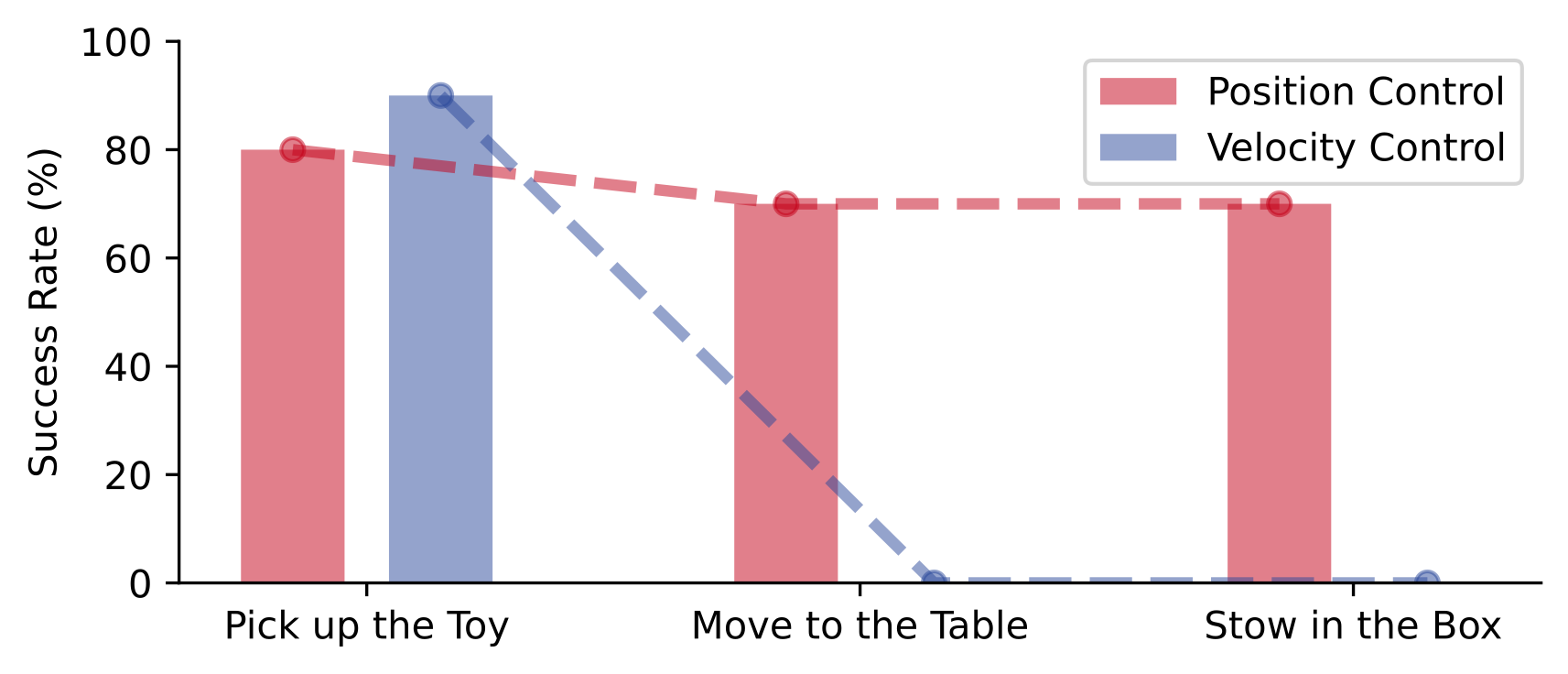}
    \caption{\textbf{Base Movement Control Ablation.} Velocity control suffers from action spikes in the training data, causing the learned policy to fail to move. Position control produces smooth trajectories and enables stable learning of base movements.}
    \label{fig:base-movement-compare}
    \vspace{-10pt}
\end{figure}

\textbf{Position vs.\ velocity control for base movement.} For the whole-body task ``Floor-to-table Pick'', fine pedal adjustments during data collection caused spikes in the base velocity distribution, preventing the policy from learning stable base motion. Switching to position control---where the policy directly predicts the target displacement from the starting point---resolved this issue. As shown in~\cref{fig:base-movement-compare}, the velocity-control policy failed to initiate movement, whereas the position-control policy learned smooth and reliable base trajectories.

\textbf{Comparison with VR-collected data.} To assess whether \tele yields data quality comparable to VR teleoperation, we conducted an additional study on ``Can Pressing'', ``Table Cleaning'', and ``Pan Sweeping''. Using a Meta Quest Pro, we collected matching datasets of 50, 200, and 200 demonstrations, respectively, and trained policies with the same codebase and training steps. As shown in~\cref{tab:teleop_compare}, policies trained on VR data and \tele data achieve similar success rates (70\% vs.\ 73\%), with only minor differences attributable to experimental variance. This parity is expected because both methods share the same control loop, and the in-the-loop pipeline keeps robot dynamics consistent between data collection and playback. In addition, we use absolute joint commands, so the impact of data noise on training is limited. Importantly, \tele attains comparable performance at substantially lower cost (\$50 vs.\ \$1,500).

\begin{table}[!htbp]
    \centering
    \small
    \setlength{\tabcolsep}{4pt}
    \renewcommand{\arraystretch}{1.1}
    \caption{Success rate comparison between policies trained on \tele and VR (Meta Quest Pro) demonstrations.}
    \label{tab:teleop_compare}
    \begin{tabular}{cccc}
        \toprule
        \textbf{Task} & \textbf{Step} & \textbf{SR (RoboPilot)} & \textbf{SR (VR)} \\
        \midrule
        \multirow{2}{*}{Can Pressing} & Locate can & 10/10 & 10/10 \\
        & Press button & 6/10 & 7/10 \\
        \multirow{3}{*}{Table Cleaning} & Grab sponge & 9/10 & 8/10 \\
        & Cleanup table & 7/10 & 6/10 \\
        & Return sponge & 7/10 & 5/10 \\
        \multirow{3}{*}{Pan Sweeping} & Lift broom & 8/10 & 8/10 \\
        & Hold dustpan & 6/10 & 7/10 \\
        & Sweep in & 5/10 & 5/10 \\
        \midrule
        Average & -- & 73\% & 70\% \\
        \bottomrule
    \end{tabular}
\end{table}

\FloatBarrier
\section{Conclusions}

This paper presents \rob, a fully open-source bimanual mobile manipulator with a total hardware budget of \$\textit{1,000--1,800}, designed to lower the hardware barrier for embodied-AI data collection and algorithm validation. A SCARA-like horizontal arm configuration eliminates counter-gravity torque demands at the joints, and dual-motor backlash elimination combined with dither-based friction compensation achieves 0.7\,mm repeatability on low-cost components. The proposed \tele teleoperation scheme employs a 26-faced marker handle that reduces tracking error by 80\%, costs only \$50, and supports fully remote operation for large-scale crowdsourced data collection. Experiments show that \tele improves data-collection efficiency by 30\% over SpaceMouse and leader-follower baselines, and imitation-learning policies trained on \tele-collected data achieve success rates comparable to those trained on VR-collected data. \rob can complete long-horizon household tasks entirely via remote teleoperation and autonomously execute complex behaviors through imitation learning. All software, CAD files, and documentation are publicly released to support low-cost research and practical deployment. Limitations remain: the robot body is relatively heavy (51\,kg) and lacks collision sensing, and vision-based teleoperation is subject to network transmission latency, limiting responsiveness for highly dynamic tasks. Future work will integrate autonomous navigation, explore lighter-weight mechanical designs, and extend the platform to service applications such as elderly care.

\bibliography{references}

@article{pfeifer2007self,
  title={Self-organization, embodiment, and biologically inspired robotics},
  author={Pfeifer, Rolf and Lungarella, Max and Iida, Fumiya},
  journal={Science},
  year={2007}
}

@inproceedings{bajracharyaDemonstratingMobileManipulation2023,
  title={Demonstrating Mobile Manipulation in the Wild: A Metrics-Driven Approach},
  author={Bajracharya, Max and Borders, James and Cheng, Richard and Helmick, Dan and Kaul, Lukas and Kruse, Dan and Leichty, John and Ma, Jeremy and Matl, Carolyn and Michel, Frank and Papazov, Chavdar and Petersen, Josh and Shankar, Krishna and Tjersland, Mark},
  booktitle={RSS},
  year={2023}
}

@inproceedings{bajracharyaMobileManipulationSystem2020,
  title={A Mobile Manipulation System for One-Shot Teaching of Complex Tasks in Homes},
  author={Bajracharya, Max and Borders, James and Helmick, Dan and Kollar, Thomas and Laskey, Michael and Leichty, John and Ma, Jeremy and Nagarajan, Umashankar and Ochiai, Akiyoshi and Petersen, Josh and Shankar, Krishna and Stone, Kevin and Takaoka, Yutaka},
  booktitle={ICRA},
  year={2020}
}

@article{black$p_0$VisionLanguageActionFlow2024,
  title={$\pi_0$: A Vision-Language-Action Flow Model for General Robot Control},
  author={$\pi_0$ Team},
  journal={arXiv:2410.24164},
  year={2024}
}

@inproceedings{chengOpenTeleVisionTeleoperationImmersive2024,
  title={Open-TeleVision: Teleoperation with Immersive Active Visual Feedback},
  author={Cheng, Xuxin and Li, Jialong and Yang, Shiqi and Yang, Ge and Wang, Xiaolong},
  booktitle={CoRL},
  year={2024}
}

@inproceedings{chiDiffusionPolicyVisuomotor2023,
  title={Diffusion Policy: Visuomotor Policy Learning via Action Diffusion},
  author={Chi, Cheng and Feng, Siyuan and Du, Yilun and Xu, Zhenjia and Cousineau, Eric and Burchfiel, Benjamin CM and Song, Shuran},
  booktitle={RSS},
  year={2023}
}

@inproceedings{chiUniversalManipulationInterface2024,
  title={Universal Manipulation Interface: In-The-Wild Robot Teaching Without In-The-Wild Robots},
  author={Chi, Cheng and Xu, Zhenjia and Pan, Chuer and Cousineau, Eric and Burchfiel, Benjamin and Feng, Siyuan and Tedrake, Russ and Song, Shuran},
  booktitle={RSS},
  year={2024}
}

@article{dingBunnyVisionProRealTimeBimanual2024,
  title={Bunny-VisionPro: Real-Time Bimanual Dexterous Teleoperation for Imitation Learning}, 
  author={Runyu Ding and Yuzhe Qin and Jiyue Zhu and Chengzhe Jia and Shiqi Yang and Ruihan Yang and Xiaojuan Qi and Xiaolong Wang},
  year={2024},
  eprint={2407.03162},
  archivePrefix={arXiv},
  primaryClass={cs.RO},
  journal={arXiv:2407.03162},
}

@inproceedings{fangAirExoLowCostExoskeletons2024,
  title={AirExo: Low-Cost Exoskeletons for Learning Whole-Arm Manipulation in the Wild},
  author={Fang, Hongjie and Fang, Hao-Shu and Wang, Yiming and Ren, Jieji and Chen, Jingjing and Zhang, Ruo and Wang, Weiming and Lu, Cewu},
  booktitle={ICRA},
  year={2024}
}

@inproceedings{fuMobileALOHALearning2024,
  title={Mobile ALOHA: Learning Bimanual Mobile Manipulation Using Low-Cost Whole-Body Teleoperation},
  author={Fu, Zipeng and Zhao, Tony Z. and Finn, Chelsea},
  booktitle={CoRL},
  year={2024}
}

@inproceedings{ghoshOctoOpenSourceGeneralist2024,
  title={Octo: An Open-Source Generalist Robot Policy},
  author={Octo team},
  booktitle={RSS},
  year={2024}
}

@inproceedings{heOmniH2OUniversalDexterous2024,
  title={OmniH2O: Universal and Dexterous Human-to-Humanoid Whole-Body Teleoperation and Learning},
  author={He, Tairan and Luo, Zhengyi and He, Xialin and Xiao, Wenli and Zhang, Chong and Zhang, Weinan and Kitani, Kris M. and Liu, Changliu and Shi, Guanya},
  booktitle={CoRL},
  year={2024}
}

@article{honerkampWholeBodyTeleoperationMobile2025,
  title={Whole-Body Teleoperation for Mobile Manipulation at Zero Added Cost},
  author={Honerkamp, Daniel and Mahesheka, Harsh and von Hartz, Jan Ole and Welschehold, Tim and Valada, Abhinav},
  journal={IEEE Robotics and Automation Letters},
  year={2025}
}

@inproceedings{iyerOPENTEACHVersatile2024,
  title={OPEN TEACH: A Versatile Teleoperation System for Robotic Manipulation},
  author={Iyer, Aadhithya and Peng, Zhuoran and Dai, Yinlong and Guzey, Irmak and Haldar, Siddhant and Chintala, Soumith and Pinto, Lerrel},
  booktitle={CoRL},
  year={2024}
}

@inproceedings{kempDesignStretchCompact2022,
  title={The Design of Stretch: A Compact, Lightweight Mobile Manipulator for Indoor Human Environments},
  author={Kemp, Charles C. and Edsinger, Aaron and Clever, Henry M. and Matulevich, Blaine},
  booktitle={ICRA},
  year={2022}
}

@inproceedings{khazatskyDROIDLargeScaleInTheWild2024,
  title = {DROID: A Large-Scale In-The-Wild Robot Manipulation Dataset},
  booktitle = {RSS},
  author = {DROID Team},
  year = {2024},
}

@inproceedings{kimOpenVLAOpenSourceVisionLanguageAction2025,
  title={OpenVLA: An Open-Source Vision-Language-Action Model},
  author={Kim, Moo Jin and Pertsch, Karl and Karamcheti, Siddharth and Xiao, Ted and Balakrishna, Ashwin and Nair, Suraj and Rafailov, Rafael and Foster, Ethan P. and Sanketi, Pannag R. and Vuong, Quan and Kollar, Thomas and Burchfiel, Benjamin and Tedrake, Russ and Sadigh, Dorsa and Levine, Sergey and Liang, Percy and Finn, Chelsea},
  booktitle={CoRL},
  year={2025}
}

@article{lenzNimbRoWinsANA2023,
  title={NimbRo Wins ANA Avatar XPRIZE Immersive Telepresence Competition: Human-Centric Evaluation and Lessons Learned},
  author={Lenz, Christian and Schwarz, Max and Rochow, Andre and P{\"a}tzold, Bastian and Memmesheimer, Raphael and Schreiber, Michael and Behnke, Sven},
  journal={International Journal of Social Robotics},
  year={2023}
}

@inproceedings{liDynamicInteractionControl2024,
  title={Dynamic Interaction Control in Legged Mobile Manipulators: A Decoupled Approach},
  author={Li, Qikai and Meng, Qinchen and Qin, Yuxing and Chen, Jiawei and Ding, Xilun and Xu, Kun},
  booktitle={ICRA},
  year={2024}
}

@inproceedings{liuDemonstratingOKRobotWhat2024,
  title={Demonstrating OK-Robot: What Really Matters in Integrating Open-Knowledge Models for Robotics},
  author={Liu, Peiqi and Orru, Yaswanth and Vakil, Jay and Paxton, Chris and Shafiullah, Nur Muhammad Mahi and Pinto, Lerrel},
  booktitle={RSS},
  year={2024}
}

@inproceedings{liuRDT1BDIFFUSIONFOUNDATION2025,
  title = {RDT-1B: A DIFFUSION FOUNDATION MODEL FOR BIMANUAL MANIPULATION},
  author = {Liu, Songming and Wu, Lingxuan and Li, Bangguo and Tan, Hengkai and Chen, Huayu and Wang, Zhengyi and Xu, Ke and Su, Hang and Zhu, Jun},
  year = {2025},
  booktitle = {ICLR},
  langid = {english}
}

@article{luoPreciseDexterousRobotic2024,
  title={Precise and Dexterous Robotic Manipulation via Human-in-the-Loop Reinforcement Learning},
  author={Luo, Jianlan and Xu, Charles and Wu, Jeffrey and Levine, Sergey},
  journal={arxiv:2410.21845},
  year={2024}
}

@inproceedings{olsonAprilTagRobustFlexible2011,
  title={AprilTag: A Robust and Flexible Visual Fiducial System},
  author={Olson, Edwin},
  booktitle={ICRA},
  year={2011}
}

@article{olssonFrictionModelsFriction1998,
  title={Friction Models and Friction Compensation},
  author={Olsson, H. and {\AA}str{\"o}m, K. J. and {Canudas de Wit}, C. and G{\"a}fvert, M. and Lischinsky, P.},
  journal={European Journal of Control},
  year={1998}
}

@inproceedings{pengRevolutionizingBatteryDisassembly2024,
  title={Revolutionizing Battery Disassembly: The Design and Implementation of a Battery Disassembly Autonomous Mobile Manipulator Robot(BEAM-1)},
  author={Peng, Yanlong and Wang, Zhigang and Zhang, Yisheng and Zhang, Shengmin and Cai, Nan and Wu, Fan and Chen, Ming},
  booktitle={IROS},
  year={2024}
}

@article{schweighoferRobustPoseEstimation2006,
  title = {Robust Pose Estimation from a Planar Target},
  author = {Schweighofer, G. and Pinz, A.},
  journal = {IEEE Transactions on Pattern Analysis and Machine Intelligence},
  year = {2006}
}

@inproceedings{setapenMARIOnETMotionAcquisition2010,
  title = {MARIOnET: Motion Acquisition for Robots through Iterative Online Evaluative Training},
  author = {Setapen, Adam and Quinlan, Michael and Stone, Peter},
  booktitle = {AAMAS},
  year = {2010}
}

@inproceedings{shawBimanualDexterityComplex2024,
  title = {Bimanual Dexterity for Complex Tasks},
  author = {Shaw, Kenneth and Li, Yulong and Yang, Jiahui and Srirama, Mohan Kumar and Liu, Ray and Xiong, Haoyu and Mendonca, Russell and Pathak, Deepak},
  booktitle = {CoRL},
  year = {2024}
}

@inproceedings{smithDesignStickbugSixArmed2024,
  title = {Design of Stickbug: A Six-Armed Precision Pollination Robot},
  author = {Smith, Trevor and Rijal, Madhav and Tatsch, Christopher and Butts, R. Michael and Beard, Jared and Cook, R. Tyler and Chu, Andy and Gross, Jason and Gu, Yu},
  booktitle = {IROS},
  year = {2024}
}

@inproceedings{spahnDemonstratingAdaptiveMobile2024,
  title = {Demonstrating Adaptive Mobile Manipulation in Retail Environments},
  author = {Spahn, Max and Pezzato, Corrado and Salmi, Chadi and Dekker, Rick and Wang, Cong and Pek, Christian and Kober, Jens and {Alonso-Mora}, Javier and Corbato, Carlos Hernandez and Wisse, Martijn},
  booktitle = {RSS},
  year = {2024}
}

@article{stantonTeleoperationHumanoidRobot2012,
  title = {Teleoperation of a Humanoid Robot Using Full-Body Motion Capture, Example Movements, and Machine Learning},
  author = {Stanton, Christopher and Bogdanovych, Anton and Ratanasena, Edward},
  journal = {Australasian Conference on Robotics and Automation},
  year = {2012}
}

@article{wuFastUMIScalableHardwareIndependent,
  title = {Fast-UMI: A Scalable and Hardware-Independent Universal Manipulation Interface},
  author = {Wu, Ziniu and Wang, Tianyu and Guan, Chuyue and Jia, Zhongjie and Liang, Shuai and Song, Haoming and Qu, Delin and Wang, Dong and Wang, Zhigang and Cao, Nieqing and Ding, Yan and Zhao, Bin and Li, Xuelong},
  journal = {arXiv:2409.19499},
  year = {2024}
}

@inproceedings{wuGELLOGeneralLowCost2023,
  title = {GELLO: A General, Low-Cost, and Intuitive Teleoperation Framework for Robot Manipulators},
  author = {Wu, Philipp and Shentu, Fred and Lin, Xingyu and Abbeel, Pieter},
  booktitle = {CoRL},
  year = {2023}
}

@inproceedings{wuTidyBotOpenSourceHolonomic2024,
  title = {TidyBot++: An Open-Source Holonomic Mobile Manipulator for Robot Learning},
  author = {Wu, Jimmy and Chong, William and Holmberg, Robert and Prasad, Aaditya and Gao, Yihuai and Khatib, Oussama and Song, Shuran and Rusinkiewicz, Szymon and Bohg, Jeannette},
  booktitle = {CoRL},
  year = {2024}
}

@inproceedings{xieCoupledActivePerception2024,
  title = {Coupled Active Perception and Manipulation Planning for a Mobile Manipulator in Precision Agriculture Applications},
  author = {Xie, Shuangyu and Hu, Chengsong and Wang, Di and Johnson, Joe and Bagavathiannan, Muthukumar and Song, Dezhen},
  booktitle = {ICRA},
  year = {2024}
}

@article{xiongAdaptiveMobileManipulation2024,
  title = {Adaptive Mobile Manipulation for Articulated Objects In the Open World},
  author = {Xiong, Haoyu and Mendonca, Russell and Shaw, Kenneth and Pathak, Deepak},
  year = {2024},
  journal = {arXiv:2401.14403}
}

@inproceedings{yangACECrossplatformVisualExoskeletons2024,
  title = {ACE: A Cross-Platform and Visual-Exoskeletons System for Low-Cost Dexterous Teleoperation},
  author = {Yang, Shiqi and Liu, Minghuan and Qin, Yuzhe and Ding, Runyu and Li, Jialong and Cheng, Xuxin and Yang, Ruihan and Yi, Sha and Wang, Xiaolong},
  booktitle = {CoRL},
  year = {2024}
}

@inproceedings{zhangLearningOpenTraverse2024,
  title = {Learning to Open and Traverse Doors with a Legged Manipulator},
  booktitle = {CoRL},
  author = {Zhang, Mike and Ma, Yuntao and Miki, Takahiro and Hutter, Marco},
  year = {2024}
}

@inproceedings{zhangNaVidVideobasedVLM2024,
  title = {NaVid: Video-Based VLM Plans the Next Step for Vision-and-Language Navigation},
  booktitle = {RSS},
  author = {Zhang, Jiazhao and Wang, Kunyu and Xu, Rongtao and Zhou, Gengze and Hong, Yicong and Fang, Xiaomeng and Wu, Qi and Zhang, Zhizheng and Wang, He},
  year = {2024}
}

@inproceedings{zhaoLearningFineGrainedBimanual2023,
  title = {Learning Fine-Grained Bimanual Manipulation with Low-Cost Hardware},
  booktitle = {RSS},
  author = {Zhao, Tony Z. and Kumar, Vikash and Levine, Sergey and Finn, Chelsea},
  year = {2023}
}

@incollection{zhouNavGPT2UnleashingNavigational2025,
  title = {NavGPT-2: Unleashing Navigational Reasoning Capability for Large Vision-Language Models},
  booktitle = {ECCV},
  author = {Zhou, Gengze and Hong, Yicong and Wang, Zun and Wang, Xin Eric and Wu, Qi},
  year = {2025}
}

@article{hartleyRotationAveraging2013,
  title = {Rotation Averaging},
  author = {Hartley, Richard and Trumpf, Jochen and Dai, Yuchao and Li, Hongdong},
  journal = {International Journal of Computer Vision},
  year = {2013},
}

@misc{pollenReachy2023,
  author = {{Pollen Robotics}},
  title = {Reachy},
  year = {2023},
  url = {https://github.com/pollen-robotics/reachy_2023},
}

@misc{palTiago2015,
  author = {{PAL Robotics}},
  title = {TIAGo},
  year = {2015},
  url = {https://pal-robotics.com/robot/tiago/},
}

@article{collinsInfinitesimalPlaneBasedPose2014,
  title = {Infinitesimal Plane-Based Pose Estimation},
  author = {Collins, Toby and Bartoli, Adrien},
  year = 2014,
  journal = {International Journal of Computer Vision},
}

@article{huangOpenPyRoA1Open2026,
  title = {OpenPyRo-A1: An Open Python-Based Low-Cost Bimanual Robot for Embodied AI},
  author = {Huang, Helong and Mower, Christopher E. and Huang, Guowei and Das, Sarthak and Dierking, Magnus and Luo, Guangyuan and Tan, Kai and Chen, Xi and Yang, Yehai and Chen, Yingbing and Zeng, Yiming and Li, Yinchuan and Zhang, Zhanpeng and Wu, Shuang and Zhang, Yingxue and Qiu, Weichao and Cao, Tongtong and Qin, Mian and Pakdamansavoji, Sajjad and Liu, Yuecheng and Zhuang, Yuzheng and Tian, Guangjian and Hao, Jianye and Wang, Jun and Bou-Ammar, Haitham and Quan, Xingyue},
  journal = {IEEE Robotics and Automation Letters},
  year = {2026},
  volume = {11},
  number = {1},
  pages = {746--753},
  doi = {10.1109/LRA.2025.3634886}
}

@misc{anjariaYORYourOwn2026,
  title = {YOR: Your Own Mobile Manipulator for Generalizable Robotics},
  author = {Anjaria, Manan H and Erciyes, Mehmet Enes and Ghatnekar, Vedant and Navarkar, Neha and Etukuru, Haritheja and Jiang, Xiaole and Patel, Kanad and Kabra, Dhawal and Wojno, Nicholas and Prayage, Radhika Ajay and Chintala, Soumith and Pinto, Lerrel and Shafiullah, Nur Muhammad Mahi and Cui, Zichen Jeff},
  year = {2026},
  url = {https://arxiv.org/abs/2602.11150}
}

@misc{wang2025xlerobot,
  title = {XLeRobot: A Practical Low-Cost Household Dual-Arm Mobile Robot Design for General Manipulation},
  author = {Wang, Gaotian and Lu, Zhuoyi and Huang, Yiyang and Liu, Yihao},
  year = {2025},
  url = {https://github.com/Vector-Wangel/XLeRobot}
}

@inproceedings{erciyesConeEOpenSource2025,
  title = {Cone-E: An Open Source Bimanual Mobile Manipulator for Generalizable Robotics},
  author = {Erciyes, M. E. and Etukuru, H. and Chintala, S. and Shafiullah, N. M. M. and Pinto, L.},
  booktitle = {RSS 2025 Workshop on Whole-Body Control and Bimanual Manipulation: Applications in Humanoids and Beyond},
  year = {2025}
}

\end{document}